\documentclass[sigconf]{acmart}

\AtBeginDocument{%
  \providecommand\BibTeX{{%
    \normalfont B\kern-0.5em{\scshape i\kern-0.25em b}\kern-0.8em\TeX}}}

\usepackage {enumitem}
\usepackage{multirow}
\setlist {nolistsep}
\usepackage{subfig}

\begin{document}
\copyrightyear{2021}
\acmYear{2021}
\setcopyright{acmcopyright}
\acmConference[MM '21]{Proceedings of the 29th ACM International Conference on Multimedia}{October 20--24, 2021}{Virtual Event, China} 
\acmBooktitle{Proceedings of the 29th ACM International Conference on Multimedia (MM '21), October 20--24, 2021, Virtual Event, China}
\acmPrice{15.00}
\acmDOI{10.1145/3474085.3475512}
\acmISBN{978-1-4503-8651-7/21/10}

\title{Privacy-Preserving Portrait Matting}

\author{Jizhizi Li$^{1*}$, Sihan Ma$^{1*}$, Jing Zhang$^{1}$, Dacheng Tao$^{2,1}$}
\thanks{*J. Li and S. Ma are co-first authors and contributed equally to this work.\\S. Ma and J. Zhang are supported by the Australian Research Council Project FL-170100117.}

\affiliation{%
  \institution{$^{1}$The University of Sydney, Sydney, Australia}
  \institution{$^{2}$JD Explore Academy, China}
  \country{}
}
\email{{jili8515, sima7436}@uni.sydney.edu.au, jing.zhang1@sydney.edu.au, dacheng.tao@gmail.com}

\renewcommand{\shortauthors}{Li and Ma, et al.}

\setlength{\abovecaptionskip}{1pt}
\setlength{\belowcaptionskip}{1pt}
\setlength{\intextsep}{1pt}
\setlength{\textfloatsep}{2pt}

\begin{abstract}
Recently, there has been an increasing concern about the privacy issue raised by using personally identifiable information in machine learning. However, previous portrait matting methods were all based on identifiable portrait images. To fill the gap, we present P3M-10k in this paper, which is the first large-scale anonymized benchmark for \textbf{P}rivacy-\textbf{P}reserving \textbf{P}ortrait \textbf{M}atting. P3M-10k consists of 10,000 high-resolution face-blurred portrait images along with high-quality alpha mattes. We systematically evaluate both trimap-free and trimap-based matting methods on P3M-10k and find that existing matting methods show different generalization capabilities when following the Privacy-Preserving Training (PPT) setting, $i.e.$, ``training on face-blurred images and testing on arbitrary images''. To devise a better trimap-free portrait matting model, we propose P3M-Net, which leverages the power of a unified framework for both semantic perception and detail matting, and specifically emphasizes the interaction between them and the encoder to facilitate the matting process. Extensive experiments on P3M-10k demonstrate that P3M-Net outperforms the state-of-the-art methods in terms of both objective metrics and subjective visual quality. Besides, it shows good generalization capacity under the PPT setting, confirming the value of P3M-10k for facilitating future research and enabling potential real-world applications. The source code and dataset are available at \href{https://github.com/JizhiziLi/P3M}{https://github.com/JizhiziLi/P3M}.
\end{abstract}

\ccsdesc[500]{Computing methodologies}
\ccsdesc[500]{Computing methodologies~Computer vision tasks}
\ccsdesc[500]{Computing methodologies~Image segmentation}

\keywords{portrait matting, deep learning, privacy-preserving, benchmark, trimap, semantic segmentation}

\maketitle

\section{Introduction}
The success of deep learning in many computer vision and multimedia areas, largely relies on large-scale of training data~\cite{zhang2020empowering}. However, for some tasks such as face recognition~\cite{masi2018deep}, human activity analysis~\cite{sun2019deep}, and speech recognition, privacy concerns about the personally identifiable information in the datasets, $e.g.$ face, gait, and voice, have attracted increasing attention recently. Unfortunately, how to alleviate the privacy concerns in data while not affecting the performance remains challenging and under-explored~\cite{yang2021study}. For instance, portrait matting, which refers to estimating the accurate foregrounds from portrait images, also involves the privacy issue, as the images usually contain identifiable faces in previous matting datasets~\cite{dim,hatt,dapm}. This issue has received more and more concerns due to the popular of virtual video meeting during the COVID-19 pandemic, since portrait matting is a key technique in this multimedia application for changing virtual background. However, we found that all the previous portrait matting methods pay less attention to the privacy issue and adopted the intact identifiable portrait images for both training and evaluation, leaving privacy-preserving portrait matting (P3M) as an open problem.

\begin{figure}[t]
    \centering
    \includegraphics[width = 1\linewidth]{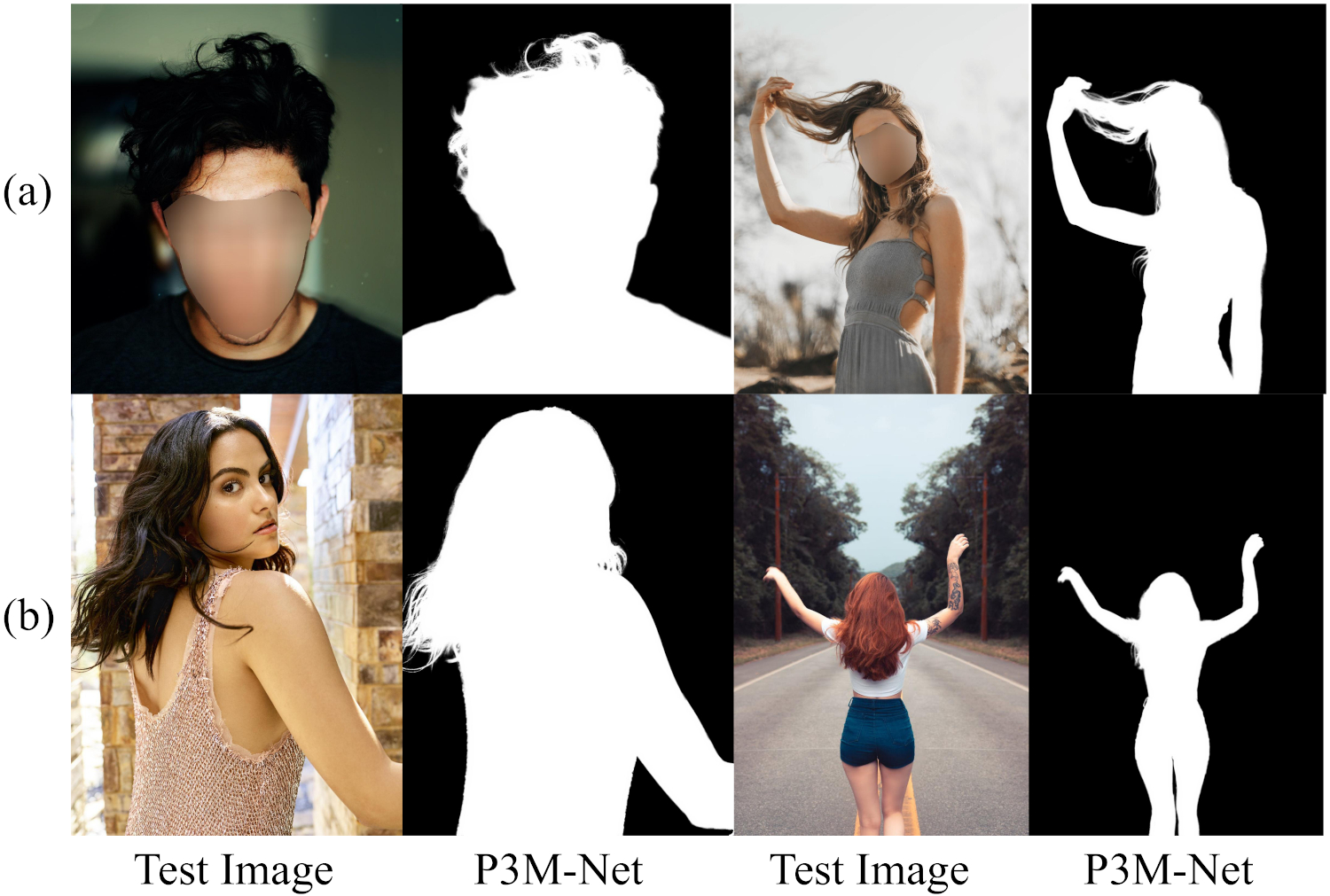}
    \caption{(a) Some anonymized portrait images from our P3M-500-P test set. (b) Some non-anonymized celebrity or back portrait images from our P3M-500-NP test set. We also provide the alpha mattes predicted by our P3M-Net, following the privacy-preserving training setting.}
    \label{fig:introduction}
\end{figure}

In this paper, we make the first attempt to fill this gap by presenting a large-scale anonymized portrait matting benchmark (P3M-10k) and investigating the impact of privacy-preserving training (PPT) on portrait matting models. P3M-10k consists of 10,000 high-resolution face-blurred portrait images where we carefully collect and filter from a huge number of images with diverse foregrounds, backgrounds and postures, along with the carefully labeled high quality ground truth alpha mattes. It surpasses the existing matting datasets~\cite{dapm,hatt,lf} in terms of diversity, volume and quality. Besides, we choose face obfuscation as the privacy protection technique to remove the identifiable face information while retaining fine details such as hairs. We split out 500 images from P3M-10k to serve as a face-blurred validation set, named P3M-500-P. Some examples are shown in Figure~\ref{fig:introduction}(a). Furthermore, to evaluate the generalization ability of matting models on normal images, which are trained following the PPT setting, $i.e.$, only using the face-blurred portrait images in P3M-10k training set, we construct a validation set with 500 images without privacy concerns, named P3M-500-NP. All the images in P3M-500-NP are either frontal images of celebrities or profile/back images without any identifiable faces. Some examples are shown in Figure~\ref{fig:introduction}(b).

It is very interesting and of significant practical meaning to see whether the privacy-preserving training will have a side impact on the matting models, since face obfuscation brings noticeable artefacts to the images which are not observed in normal portrait images. We notice that a contemporary work \cite{yang2021study} has shown empirical evidences that face obfuscation only has minor side impact on object detection and recognition models. However, in the context of portrait matting, where the pixel-wise alpha matte (a soft mask) with fine details is expected to be estimated from a high-resolution portrait image, the impact remains unclear.

In this paper, we systematically evaluate both trimap-based and trimap-free matting methods on the unmodified version and face-blurred version of P3M-10k, and provide our insight and analysis about the impact. Specifically, we found that for trimap-based matting, where the trimap is used as an auxiliary input, face obfuscation shows little impact on the matting models, $i.e.$, a slight performance change of models following the PPT setting. As for trimap-free matting which involves two sub-tasks: foreground segmentation and detail matting, we found that the methods using a multi-task framework that explicitly model and jointly optimize both tasks~\cite{gfm,hatt} are able to mitigate the impact of face obfuscation to an acceptable level (2\% to 5\%). Besides, the matting methods that solve the problem through sequential segmentation and matting \cite{shm,dapm}, show a significant performance drop, since face obfuscation leads to segmentation errors that will mislead the subsequent matting model. Other methods that involve several stages of networks to progressively refine the alpha mattes from coarse to fine, shows to be less affected by face obfuscation but still observe a performance drop due to the lack of explicit semantic guidance. Meanwhile, these methods are a little awkward due to the tedious training process.

Based on the above observations, we propose a novel automatic portrait matting network named P3M-Net, which is able to serve as a strong trimap-free matting baseline for the P3M task (See the results in Figure~\ref{fig:introduction}). Technically, we also adopt a multi-task framework like \cite{gfm,hatt} as our basic structure, which learns common visual features through a sharing encoder and task-aware features through a segmentation decoder and a matting decoder. In contrast to previous methods \cite{gfm,hatt}, we specifically emphasize the interaction between two decoders and those between encoder and decoders. To this end, we 
devise a Tripartite-Feature Integration (TFI) module for each matting decoder block to effectively fuse the matting decoder features from the previous block, semantic features from the segmentation decoder block, and base visual features from the encoder block. Besides, we devise a Deep Bipartite-Feature Integration (dBFI) module and a Shallow Bipartite-Feature Integration (sBFI) module to leverage deep features with high-level semantics and shallow features with fine details for improving the segmentation decoder and matting decoder, respectively. Experiments on the P3M-10k benchmark demonstrate that P3M-Net achieves a neglectable performance drop under the PPT setting and outperforms all the previous trimap-free matting methods by a large margin. 

To sum up, the contributions of this paper are three-fold. \textbf{First}, to the best of our knowledge, we are the first to study the problem of privacy-preserving portrait matting and establish the largest privacy-preserving portrait matting dataset P3M-10k, which can serve as benchmark for P3M. \textbf{Second}, we systematically investigate the impact of face obfuscation on both trimap-based and trimap-free matting models under the privacy-preserving training setting and provide insights about the evaluation protocol, performance analysis, and model design. \textbf{Third}, We propose a novel trimap-free portrait matting network named P3M-Net that follows a multi-task framework and specifically focuses on the interactions between encoders and decoders. P3M-Net demonstrates its value for privacy-preserving portrait matting and can serve as a strong baseline.


\section{Related Work}
\subsection{Image Matting}
Image matting is a typical ill-posed problem to estimate the foreground, background, and alpha matte from a single image. Specifically, portrait matting refers to a specific image matting task where the input image is a portrait. From the perspective of input, image matting can be divided into two categories, $i.e.$, trimap-based methods and trimap-free methods. Trimap-based matting methods use a user defined trimap, $i.e.$, a 3-class segmentation map, as an auxiliary input, which provides explicit guidance on the transition area. Previous methods include affinity-based methods~\cite{levin2007closed,aksoy2018semantic}, sampling-based methods~\cite{he2011global,shahrian2013improving}, and deep learning based methods~\cite{lu2019indices,hou2019context}. Besides, 
there are other methods used different auxiliary inputs, $e.g.$, a background image~\cite{backgroundmatting,backgroundmattingv2}, or a coarse map~\cite{yu2021mask}.

To enable automatic (portrait) image matting, recent works~\cite{shm,lf,hatt,gfm,aim} tried to estimate the alpha matte directly from a single image without using any auxiliary input, also known as trimap-free methods. For example, DAPM~\cite{dapm} and SHM~\cite{shm} tackled the task by separating it into two sequential stages, $i.e.$, segmentation and matting. However, the semantic error produced in the first stage will mislead the matting stage and can not be corrected. LF~\cite{lf} and SHMC~\cite{shmc} solved the problem by generating coarse alpha matte first and then refining it. Besides of the tedious training process, these methods suffer from ambiguous boundaries due to the lack of explicit semantic guidance. HATT~\cite{hatt} and GFM~\cite{gfm} proposed to model both the segmentation and matting tasks in a unified multi-task framework, where a sharing encoder was used to learn base visual features and two individual decoders are used to learn task-relevant features. However, HATT~\cite{hatt} lacks explicit supervision on the global guidance while GFM~\cite{gfm} lacks modeling the interactions between both tasks. By contrast, we propose a novel model named P3M-Net, which is also based on the multi-task framework but specifically focuses on modeling the interactions between encoders and decoders. Besides, we comprehensively investigate their performance under the PPT setting on P3M-10k and provide some useful insights on their model structures. 

\subsection{Privacy Issues in Visual Tasks}
There are two kinds of privacy issues in visual tasks, $i.e.$, private data protection and private content protection in public academic datasets. For the former, there are concerns of information leak caused by insecure data transferring and membership inference attacks to the trained models~\cite{shokri2017membership,hisamoto2020membership,fredrikson2015model,carlini2019secret}. Privacy-preserving machine learning (PPML) aims to solve these problems by homonorphic encryption~\cite{erkin2009privacy,yonetani2017privacy} and differential privacy algorithms~\cite{NEURIPS2020_fc4ddc15,abadi2016deep}.

For public academic datasets, there is no concern for information leak, thus PPML is no longer needed. But, there still exists privacy breach incurred by exposure of personally identifiable information, $e.g.$ faces, addresses. It is a common problem in the benchmark datasets for many visual tasks, $e.g.$ object recognition and semantic segmentation. Recently a contemporary work \cite{yang2021study} has shown empirical evidences that face obfuscation, as an effective data anonymization technique, only has minor side impact on object detection and recognition. However, since portrait matting requires to estimate a pixel-wise soft mask (alpha matte) for a high-resolution portrait image, the impact  remains unclear.

\subsection{Privacy-Preserving Methods}

Normally, to protect the private information in the public images, a common method is to capture or process the data in special low quality conditions~\cite{dai2015towards,butler2015privacy}. For example, Dai $et al$. captured the anonymized video data in extremely low resolution to avoid the leak of personally identifiable information such as frontal faces \cite{dai2015towards}. Another way is to add empirical obfuscations~\cite{uittenbogaard2019privacy,caesar2020nuscenes,frome2009large}, such as blurring and mosaicing at certain regions. Yang $et al$. used face blurring to obfuscate the faces in the ImageNet dataset \cite{yang2021study}. Caesar $et al$. detected and blurred the license plates in nuScenes dataset to avoid privacy concerns~\cite{caesar2020nuscenes}. For the portrait matting task, all previous benchmarks or methods pay little attention to the privacy issue. By contrast, we make the first attempt to construct a large-scale anonymized dataset for privacy-preserving portrait matting named P3M-10k. Specifically, we use face obfuscation as the privacy-preserving strategy to anonymize the identities of all images. Intuitively, the anonymized images with blurred faces may degenerate the performance of matting models due to the domain gap between anonymized training images and normal test images. In this paper, we make the first attempt to investigate the impact of face obfuscation on portrait matting under the PPT setting and identify that it has negligible impact on trimap-based matting methods but has different impact on trimap-free matting methods, depending on their model structure.

\subsection{Matting Datasets}
Existing matting datasets either contain only a small number of high-quality images and annotations, or the images and annotations are in low-quality. For example, the online benchmark alphamatting~\cite{TUW-180666} only provides 27 high-resolution training images and 8 test images. None of them is portrait image. Composition-1k~\cite{dim}, the most commonly used dataset, contains 431 foregrounds for training and 20 foregrounds for testing. However, many of them are consecutive video frames, making it less diverse. GFM~\cite{gfm} provides 2,000 high-resolution natural images with alpha mattes, but they are all animal images. With respect to portrait image matting dataset, DAPM~\cite{dapm} provided a large dataset of 2,000 low-resolution portrait images with alpha mattes generated by KNN matting~\cite{chen2013knn} and closed form matting~\cite{levin2007closed}, whose quality is limited. Late fusion~\cite{lf} built a human image matting dataset by combining 228 portrait images from Internet and 211 human images in Composition-1k. Distinction-646~\cite{hatt} is a dataset containing 364 human images but with only foregrounds provided. There are also some large-scale portrait datasets, $e.g.$, SHM~\cite{shm}, SHMC~\cite{shmc}, and background matting~\cite{backgroundmatting, backgroundmattingv2}, which are unfortunately not public. Most importantly, no privacy preserving method is used to anonymize the images in the aforementioned datasets, making all the frontal faces exposed. By contrast, we establish the first large-scale matting dataset with 10,000 high-resolution portrait images with high-quality alpha mattes and anonymize all images using face obfuscation.


\begin{table*}[htb]

\begin{center}
\resizebox{\textwidth}{!}{
\begin{tabular}{c|c|cccccccc}
\hline
Test set & Metric & Closed~\cite{levin2007closed} & IFM~\cite{aksoy2017designing} & KNN~\cite{chen2013knn} & Compre~\cite{shahrian2013improving} & Robust~\cite{wang2007optimized} & Learning~\cite{zheng2009learning} & Global~\cite{he2011global} & Shared~\cite{gastal2010shared}\\
\hline
\multirow{3}{*}{B} & SAD & 9.5750 & 10.887 & 15.378 & 8.3208 & 9.3321 & 10.248 & 9.6157 & 10.553 \\
 & MSE & 0.0214 & 0.0326 & 0.0511 & 0.0194 & 0.0214 & 0.0238 & 0.0242 & 0.0285 \\
 & MAD & 0.0693 & 0.0760 & 0.1087 & 0.0602 & 0.0674 & 0.0737 & 0.0708 & 0.0774 \\
 \hline
\multirow{3}{*}{N} & SAD & 9.4812 & 10.793 & 15.366 & 8.2295 & 9.2486 & 10.151 & 9.4908 & 10.386\\
 & MSE & 0.0210 & 0.0318 & 0.0506 & 0.0191 & 0.0211 & 0.0236 & 0.0236 & 0.0277\\
 & MAD & 0.0686 & 0.0748 & 0.1078 & 0.0595 & 0.0668 & 0.0729 & 0.0698 & 0.0760\\
 \hline
\end{tabular}}
\end{center}
\caption{Results of trimap-based traditional methods on the blurred images (``B'') and normal images (``N'') in P3M-500-P.}
\label{tab:benchmark_trimap_based_traditional}
\end{table*}

\begin{table*}[htb]
\begin{center}
\resizebox{\textwidth}{!}{
\begin{tabular}{c|ccc|ccc|ccc|ccc}
\hline
Setting & \multicolumn{3}{c}{B:B} & \multicolumn{3}{|c}{B:N} & \multicolumn{3}{|c}{N:B} & \multicolumn{3}{|c}{N:N} \\
\hline
Method & SAD & MSE  & MAD & SAD & MSE & MAD & SAD & MSE & MAD & SAD & MSE & MAD\\
\hline
DIM~\cite{dim} & 4.8906 & 0.0115 & 0.0342 & 4.8940 & 0.0116 & 0.0342 & 4.8050 & 0.0116 & 0.0334 & 4.7941 & 0.0116 & 0.0334 \\
AlphaGAN~\cite{bmvcLutzAS18} & 5.2669 & 0.0112 & 0.0373 & 5.2367 & 0.0112 & 0.037 & 5.7060 & 0.0120 & 0.0412 & 5.6696 & 0.0119 & 0.0408\\
GCA~\cite{li2020natural} & 4.3593 & 0.0088 & 0.0307 & 4.3469 & 0.0089 & 0.0306 & 4.4068 & 0.0089 & 0.0310 & 4.4002 & 0.0089 & 0.0310 \\
IndexNet~\cite{lu2019indices} & 5.1959 & 0.0156 & 0.0368 & 5.2188 & 0.0158 & 0.0370 & 5.8267 & 0.0202 & 0.0420 & 5.8509 & 0.0204 & 0.0422 \\
FBA~\cite{forte2020fbamatting} & 4.1330 & 0.0088 & 0.0289 & 4.1267 & 0.0088 & 0.0289 & 4.1666 & 0.0086 & 0.0291 & 4.1544 & 0.0086 & 0.0290 \\
 \hline
\end{tabular}}
\end{center}
\caption{Results of trimap-based deep learning methods on P3M-500-P. ``B'' denotes the blurred images while ``N'' denotes the normal images. ``B:N'' denotes training on blurred images while testing on normal images, vice versa.
}
\label{tab:benchmark_trimap_based_dl}
\end{table*}

\begin{table*}[htb]
\begin{center}
\resizebox{\textwidth}{!}{
\begin{tabular}{c|ccc|ccc|ccc|ccc}
\hline
Setting & \multicolumn{3}{c}{B:B} & \multicolumn{3}{|c}{B:N} & \multicolumn{3}{|c}{N:B} & \multicolumn{3}{|c}{N:N} \\
\hline
Method & SAD & MSE  & MAD & SAD & MSE & MAD & SAD & MSE & MAD & SAD & MSE & MAD\\
\hline
SHM~\cite{shm} & 21.56 & 0.0100 & 0.0125 & 24.33& 0.0116& 0.0140& 23.91& 0.0115 & 0.0139 & 17.13 & 0.0075& 0.0099\\
LF~\cite{lf} & 42.95 & 0.0191 & 0.0250& 30.84& 0.0129& 0.0178 &41.01 & 0.0174 & 0.0240 & 31.22 &0.0123& 0.0181\\
HATT~\cite{hatt} &25.99  &0.0054  &0.0152  &26.5 &0.0055 &0.0155 &35.02 &0.0103  &0.0204  &22.93  &0.0040 &0.0133 \\
GFM~\cite{gfm} & 13.20 & 0.0050 & 0.0080 & 13.08& 0.0050& 0.0080& 13.54  &0.0048  &0.0078  & 10.73& 0.0033& 0.0063\\
 \hline
 BASIC & 15.13 & 0.0058 & 0.0088 & 15.52& 0.0060& 0.0090& 24.38& 0.0109 &0.0141  & 14.52 &0.0054 &0.0085 \\
 P3M-Net (Ours) & 8.73 & 0.0026 & 0.0051 & 9.22& 0.0028& 0.0053& 11.22& 0.0040 & 0.0065 & 9.06 &0.0028 &0.0053 \\
 
 \hline
\end{tabular}}
\end{center}
\caption{Results of trimap-free methods on P3M-500-P. Please refer to Table~\ref{tab:benchmark_trimap_based_dl} for the meaning of different symbols.}
\label{tab:benchmark_e2e}
\end{table*}

\begin{figure}
    \centering
    \includegraphics[width=\linewidth]{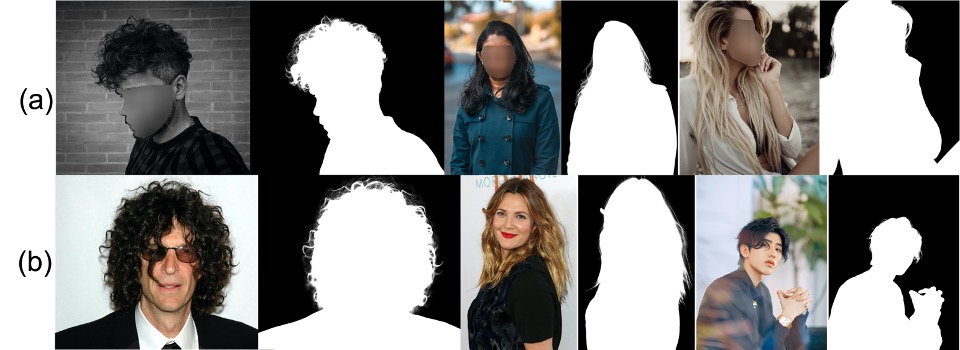}
    \caption{(a) Samples from the P3M-10k training set and P3M-500-P test set. (b) Samples from P3M-500-NP test set.}
    \label{fig:data_p3m-10k}
\end{figure}
\section{A Benchmark for P3M and Beyond}
Privacy-preserving portrait matting (P3M) is an important and meaningful topic due to the increasing privacy concerns. In this section, we first define this problem, then establish a large-scale anonymized portrait matting dataset P3M-10k to serve as a benchmark for P3M. A systematic evaluation of the existing trimap-based and trimap-free matting methods on P3M-10k is conducted to investigate the impact of the privacy-preserving training (PPT) setting on different matting models and gain some useful insights.

\subsection{PPT Setting and P3M-10k Dataset}
\subsubsection{PPT Setting}
Due to the privacy concern, we propose the \textbf{p}rivacy-\textbf{p}reserving \textbf{t}raining (PPT) setting in portrait matting, $i.e.$, training on privacy-preserved images ($e.g.$, processed by face obfuscation) and testing on arbitrary images with or without privacy content. As an initial step towards privacy-preserving portrait matting problem, we only define the identifiable faces in frontal and some profile portrait images as the private content in this work. Intuitively, PPT setting is challenging since face obfuscation brings noticeable artefacts to the images which are not observed in normal portrait images, $i.e.$, probably resulting in a domain gap between training and testing. Therefore, it is very interesting and of significant practical meaning to see whether the PPT setting will have a side impact on the matting models.

\subsubsection{P3M-10k Dataset}

To answer the above question, we establish the first large-scale privacy-preserving portrait matting benchmark named P3M-10k. It contains 10,000 anonymized high-resolution portrait images by face obfuscation along with high-quality ground truth alpha mattes. Specifically, we carefully collect, filter, and annotate about 10,000 high-resolution images from the Internet with free use license. There are 9,421 images in the training set and 500 images in the test set, denoted as P3M-500-P. In addition, we also collect and annotate another 500 public celebrity images from the Internet without face obfuscation, to evaluate the performance of matting models under the PPT setting on normal portrait images. Some examples are shown in Figure~\ref{fig:data_p3m-10k}.

Our P3M-10k outperforms existing matting datasets in terms of dataset volume, image diversity, privacy preserving, and providing natural images instead of composited ones. The diversity is not only shown in foreground, $e.g.$, half and full body, frontal, profile, and back portrait, different genders, races, and ages, $etc$., but also in background. Images in P3M-10k are captured in different indoor and outdoor environments with various illumination conditions. Some examples are shown in Figure~\ref{fig:data_p3m-10k}. In addition, we argue that large volume and high diversity of P3M-10k enable models to train on the natural images without the need of image composition. Image composition using low-resolution background images is a common practice in previous works \cite{dim, hatt} to increase data diversity due to the small dataset volume. However, there are obvious composition artefacts in the composition images due to the discrepancy of foreground and background images in noise, resolution, and illumination. By contrast, the background in the natural images are compatible with the foreground since they are captured from the same scene. The composition artefacts may have a side impact on the generalization ability of matting models as shown in \cite{gfm}. We leave it as the future work to systematically investigate this problem and only focus on the PPT setting in this paper. 

\begin{figure*}
    \centering
    \includegraphics[width=\linewidth]{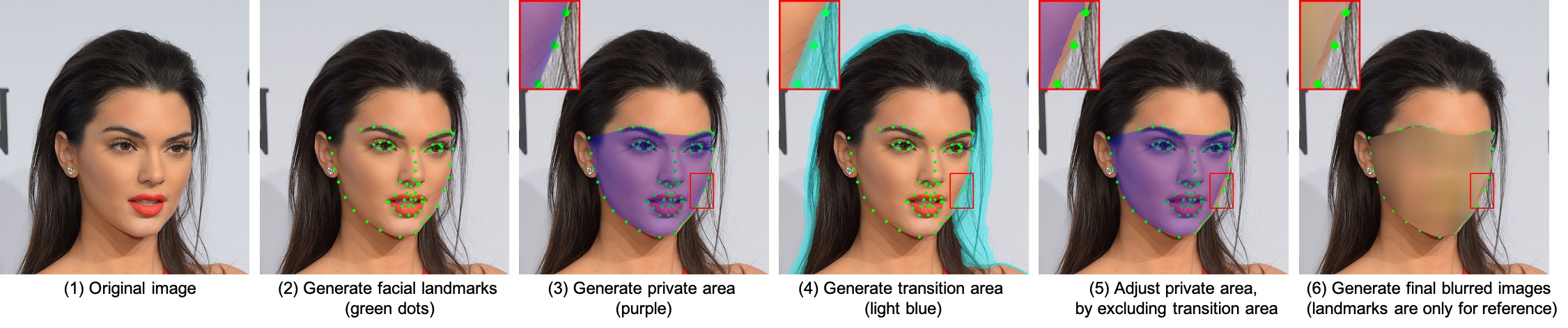}
    \caption{Illustration of the face blurring process.}
    \label{fig:face_obfuscation}
\end{figure*}

\subsection{Privacy-Preserving Method in P3M-10k}
We propose to use blurring to obfuscate the identifiable faces. Instead of using a face detector to obtain the bounding box of face and blurring it accordingly as in \cite{yang2021study}, we adopt facial landmark detectors~\cite{bulat2017far, zhang2021towards} to obtain the face mask. It is because different from the classification and detection tasks in \cite{yang2021study}, which may not be sensitive to the blurry boundaries, portrait matting requires to estimate the foreground alpha matte with clear boundaries, including the transition areas of face such as cheek and hair. As shown in Figure~\ref{fig:face_obfuscation}, after obtaining the landmarks, a pixel-level face mask is automatically generated along the cheek and eyebrow landmarks in step (3). Then, we exclude the transition area shown in step (4) and generate an adjusted face mask at step (5). Finally, we use Gaussian blur to obfuscate the identifiable faces in the mask and the final result is shown in step (6). Note that for those images with failure landmark detection, we manually annotate the face mask.

\subsection{Benchmark Setup}
\subsubsection{Methods}
We evaluate both trimap-based and trimap-free matting methods. The full list of methods are shown in Table~\ref{tab:benchmark_trimap_based_traditional},~\ref{tab:benchmark_trimap_based_dl},~\ref{tab:benchmark_e2e}.

\subsubsection{Evaluation Metrics} 
We use the common metrics including MSE, SAD, and MAD for evaluation. For trimap-based methods, the metrics are only calculated over the transition area, while for trimap-free methods, they are calculated over the whole image.

\subsubsection{Training and Evaluation Protocols}
Four kinds of training and evaluation protocols are proposed, including ``trained on blurred images, test on blurred ones (B:B)'', ``trained on blurred images and test on normal ones (B:N)'', ``trained on normal images and test on blurred ones (N:B)'', and ``trained on normal images and test on normal ones (N:N)''. The first two protocols correspond to the proposed PPT settings. All the methods are trained using the normal or blurred images in the P3M-10k training set and evaluated on P3M-500-P test set. The only difference between blurred and normal images is whether or not face obfuscation is applied.


\subsection{Study on The Impact of PPT}

\subsubsection{Impact on Trimap-based Traditional Methods}
As in Table~\ref{tab:benchmark_trimap_based_traditional}, trimap-based traditional methods show neglectable performance variance under different training and evaluation protocols, indicating that PPT setting brings little impact on these methods. This observation is reasonable, since traditional methods mainly make prediction based on local pixels in the transition area with no blurring, although a few of sampled neighboring pixels may be blurred. 

\subsubsection{Impact on Trimap-based Deep Learning Methods}
Similar to traditional trimap-based methods, deep learning methods also show very minor changes across different settings, as shown in Table~\ref{tab:benchmark_trimap_based_dl}. This is because trimap-based deep learning methods use the ground truth trimap as an auxiliary input and focus on estimating the alpha matte of the transition area, probably guiding the model to pay less attention to the blurred areas. In addition, there are also some observations opposed to intuition. When testing on normal images, models trained on the normal training images surprisingly fall behind of those trained on the blurred ones. For instance, the SAD of IndexNet on ``N:N'' is 0.6 higher than the score on ``B:N''. Similar results can also be found for AlphaGAN, GCA in Table~\ref{tab:benchmark_trimap_based_dl}.
We suspect that the blurred pixels near the transition area may serve as a random noise during the training process, which makes the model more robust and leads to a better generalization. 

\subsubsection{Impact on Trimap-free Methods}
Different from trimap-based methods, trimap-free methods show significant performance changes under four protocols. The results are shown in Table~\ref{tab:benchmark_e2e}. \textbf{First}, we start with the test set of normal images by comparing results in the ``B:N'' and ``N:N'' tracks. Models trained on normal training images (N:N) usually outperform those using the blurred ones (B:N), $e.g.$, from 24.33 to 17.13 at SAD for SHM. This observation makes sense since there is a domain gap between blurred images and normal ones due to face obfuscation. By comparison, we found trimap-free methods show different generalization ability in dealing with this domain gap. For example, SHM is the worst with a large drop of 7 in SAD, while HATT and GFM only show a drop less than 3 in SAD. We suspect that an end-to-end multi-task framework may probably mitigate the domain gap issue via joint optimization. By contrast, two-stage methods such as SHM may produce segmentation errors, which can mislead the following matting stage and can not be corrected. To validate this hypothesis, we devise a baseline model called ``BASIC'' by adopting a similar multi-task framework like HATT and GFM but removing the bells and whistles, $i.e.$, only using a sharing encoder and two individual decoders. As shown in Table~\ref{tab:benchmark_e2e}, the small performance drop (less than 1 in SAD) proves its superiority in overcoming domain gap and supports our hypothesis.

\textbf{Second}, we focus on the test set of blurred images by comparing results in the ``B:B'' and ``N:B'' tracks. The performance drop in most methods, $e.g.$, 9.03 in SAD for HATT, proves that without seeing the blurred pattern during training, models cannot generalize well to the face-blurred images. It implies the value of the blurred training set in P3M-10k for training models that will be deployed in privacy-preserving scenarios, where faces may be blurred.

\textbf{Third}, we fix the training set to be the blurred one. By comparing the ``B:B'' and ``B:N'' tracks, we observe similar performance of most methods under these two settings, $e.g.$, SADs of HATT are 25.99 and 26.5 on the blurred test set and normal one.
These results imply that we can use the performance on the blurred test images in P3M-500-P as a bold indicator of that on the normal ones.

\label{sec:ppt}


\begin{figure*}[t]
    \centering
    \includegraphics[width = \linewidth]{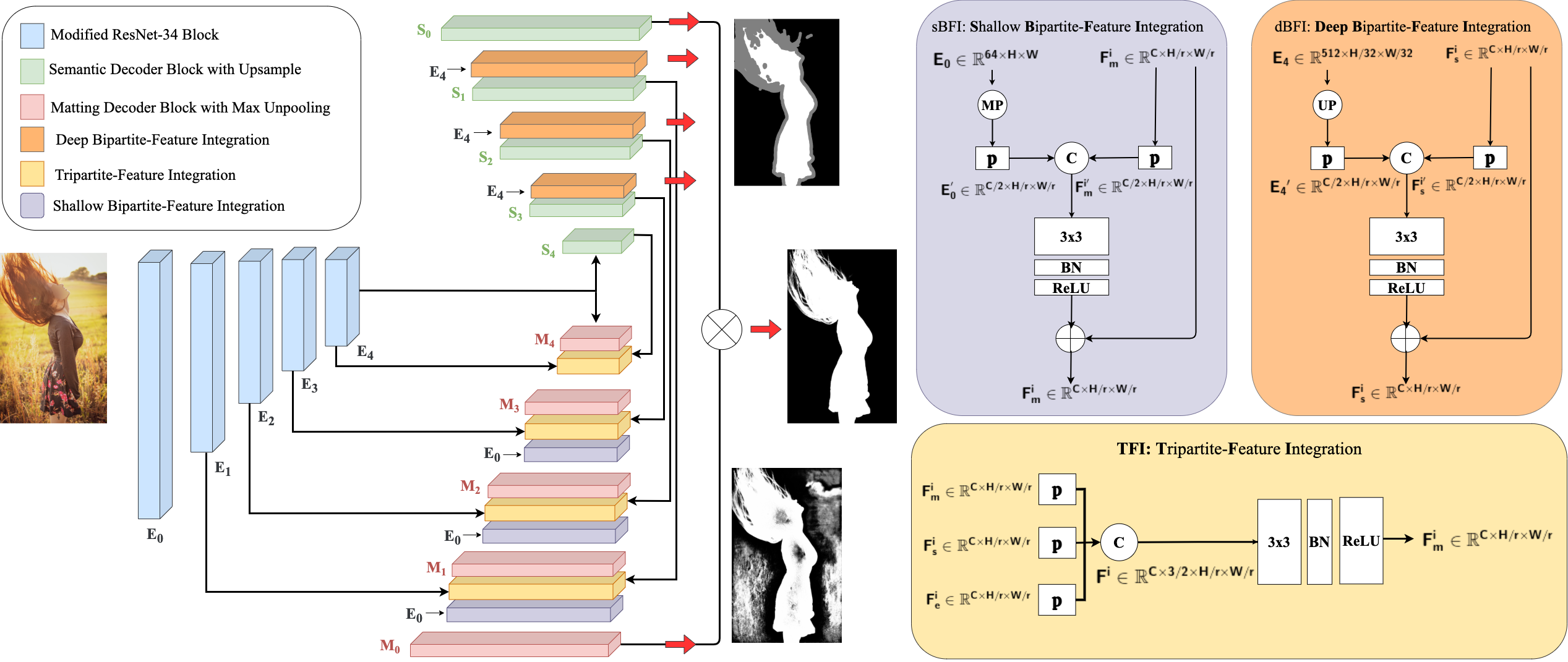}
    \caption{Diagram of the proposed P3M-Net structure. It adopts a multi-task framework, which consists of a sharing encoder, a segmentation decoder, and a matting decoder. Specifically, a TFI module, a dBFI module, and a sBFI module are devised to model different interactions among the encoder and the two decoders. Red arrows denote the network's outputs.}
    \label{fig:network}
\end{figure*}
\section{A Strong Baseline for P3M}


\subsection{A Multi-task Framework}
As discussed in Section~\ref{sec:ppt}, trimap-free matting models benefit from explicitly modeling both semantic segmentation and detail matting and jointly optimizing them in an end-to-end multi-task framework. Therefore, we follow GFM~\cite{gfm} to adopt the multi-task framework, where base visual features are learned from a sharing encoder and task-relevant features from individual decoders, $i.e.$, semantic decoder and matting decoder, respectively. For the sharing encoder, we choose a modified version of ResNet-34~\cite{he2016deep} with max pooling layers to serve as our light-weight backbone as in AIM~\cite{aim}. We also keep the indices of the max pooling operation to reserve the details and used in the unpooling layers in the matting decoder. Both semantic decoder and matting decoder have five blocks, each of which contains three convolution layers. We then choose different upsampling operations to suit each task. We use bilinear interpolation in the semantic decoder for simplicity and use max unpooling operation with the indices from corresponding encoder blocks in the matting decoder to learn features for fine details.

\subsection{TFI: Tripartite-Feature Integration}
\label{sec:tim}
Most of the previous matting methods either model the interaction between encoder and decoder such as the U-Net~\cite{unet} style structure in \cite{gfm} or model the interaction between two decoders such as the attention module in~\cite{hatt}. In this paper, we comprehensively model all the interactions between the sharing encoder and two decoders, $i.e.$, 1) a tripartite-feature integration (TFI) module in each matting decoder block to model the interaction between encoder, segmentation decoder, and the matting decoder; 2) a deep bipartite-feature integration (dBFI) module to model the interaction between the encoder and segmentation decoder; and 3) a shallow bipartite-feature integration (sBFI) module to model the interaction between the encoder and matting decoder.

Specifically, for each \textbf{TFI}, it has three inputs, $i.e.$, the feature map of the previous matting decoder block $\mathbf{F_m^i} \in \mathbb{R}^{C\times H/r \times W/r}$, the feature map from same level semantic decoder block $\mathbf{F_s^i} \in \mathbb{R}^{C\times H/r \times W/r}$, and the feature map from the symmetrical encoder block $\mathbf{F_e^i} \in \mathbb{R}^{C\times H/r \times W/r}$, where $i\in \{1,2,3,4\}$ stands for the block index, $r$ stands for the downsample ratio of the feature map compared to the input size, and $r=2^i$. For each feature map, we use an $1\times1$ convolutional projection layer $\mathcal{P}$ for further embedding and channel reduction. The output of $\mathcal{P}$ for each feature map is $\mathbf{F^i} \in \mathbb{R}^{C/2\times H/r \times W/r}$. We then concatenate the three embedded feature maps and feed them into a convolutional block $\mathcal{C}$ containing a
$3\times3$ convolutional layer, a batch normalization layer, and a ReLU layer. As shown in Eq.~\ref{equa:tim}, the output feature is $\mathbf{F_m^i}\in\mathbb{R}^{C\times H/r \times W/r}$:
\begin{equation}
\mathbf{F_m^i} = \mathcal{C}(Concat(\mathcal{P}(\mathbf{F^i_m}),\mathcal{P}(\mathbf{F^i_s}),\mathcal{P}(\mathbf{F^i_e}))).
\label{equa:tim}
\end{equation}

\subsection{sBFI: Shallow Bipartite-Feature Integration}
The matting task requires to distinguish fine foreground details from the background. Therefore, the features in the shallow layers in the encoder may be useful since they contain abundant structural detail features. To leverage them to improve the matting decoder, we propose the shallow bipartite-feature integration (sBFI) module.

As shown in Figure~\ref{fig:network}, we use the feature map $\mathbf{E_0}\in \mathbb{R}^{64\times H \times W}$ in the first encoder block as a guidance to refine the output feature map $\mathbf{F_m^i}\in\mathbb{R}^{C\times H/r \times W/r}$ from the previous matting decoder block since $\mathbf{E_0}$ contains many details and local structural information. Here, $i\in \{1,2,3\}$ stands for the layer index, $r$ stands for the downsample ratio of the feature map compared to the input size, and $r=2^i$. Since $\mathbf{E_0}$ and $\mathbf{F_m^i}$ are with different resolution, we first adopt max pooling $MP$ with a ratio $r$ on $\mathbf{E_0}$ to generate a low-resolution feature map $\mathbf{E_0^{'}}\in \mathbb{R}^{64\times H/r \times W/r}$. We then feed both $\mathbf{E_0^{'}}$ and $\mathbf{F_m^i}$ to two projection layers $\mathcal{P}$ implemented by $1\times1$ convolution layers for further embedding and channel reduction, $i.e.$, from $C$ to $C/2$. Finally, the two feature maps are concatenated and and fed into a convolutional block $\mathcal{C}$ containing a $3\times3$ convolutional layer, a bacth normalization layer, and a ReLu layer. As shown in Eq.~\ref{equa:sgr}, we adopt the residual learning idea by adding the output feature map back to the input matting decoder feature map $\mathbf{F_m^i}$:
\begin{equation}
\mathbf{F_m^i} = \mathcal{C}(Concat(\mathcal{P}(\mathcal{MP}(\mathbf{E_0})),\mathcal{P}(\mathbf{F^i_m})))+\mathbf{F_m^i}.
\label{equa:sgr}
\end{equation}
In this way, sBFI helps the matting decoder block to focus on the fine details guided by $\mathbf{E_0}$.

\subsection{dBFI: Deep Bipartite-Feature Integration}
Same as sBFI, features in the encoder can also provide valuable guidance to the segmentation decoder. In contrast to sBFI, we chose the feature map $\mathbf{E_4}\in \mathbb{R}^{512\times H/32 \times W/32}$ from the last encoder block, since it encodes abundant global semantics. 

Specifically, we devise the deep bipartite-feature integration (dBFI) module to fuse it with the feature map $\mathbf{F_s^i}\in\mathbb{R}^{C\times H/r \times W/r}$ from the $i$th segmentation decoder block to improve the feature representation ability for the high-level semantic segmentation task. Here, $i\in \{1,2,3\}$. Note that since $\mathbf{E_4}$ is in low-resolution, we use a upsampling operation $UP$ with a ratio $32/r$ on $\mathbf{E_4}$ to generate $\mathbf{E_4^{'}}\in \mathbb{R}^{512\times H/r \times W/r}$. we then feed both $\mathbf{E_4^{'}}$ and $\mathbf{F_s^i}$ into two projection layers $\mathcal{P}$, concatenated together, and fed into a convolutional block $\mathcal{C}$. We adopt the identical structures for $\mathcal{P}$ and $\mathcal{C}$ as those in sBFI. Similarly, this process can be described as: 
\begin{equation}
\mathbf{F_s^i} = \mathcal{C}(Concat(\mathcal{P}(\mathcal{UP}(\mathbf{E_4})),\mathcal{P}(\mathbf{F^i_s})))+\mathbf{F_s^i}
\label{equa:dgr}
\end{equation}
Note that we reuse the symbols of $\mathcal{C}$ and $\mathcal{P}$ in Eq.~\ref{equa:tim}, Eq.~\ref{equa:sgr}, and Eq.~\ref{equa:dgr} for simplicity, although each of them denotes a specific layer (block) in TFI, sBFI, and dFI, respectively.

\subsection{Training Objective}
For the segmentation task, we leverage the deep supervision idea and add side loss on the segmentation decoder to stable and improve the training performance. Specifically, we use a $3x3$ convolution layer and an upsample operation on each output feature map $\mathbf{F_s^i}$ from dBFI blocks, to predict the segmentation map with 3 channels and in the same resolution as input, denoting as ${P_s^{i}}\in \mathbb{R}^{3\times H \times W}$. 
We then calculate the cross-entropy loss $\mathcal{L}_{CE}^i$ between ${P_s^{i}}$ and the ground truth trimap label $G \in \mathbb{R}^{3\times H \times W}$ ($i.e.$, the one-shot representation) defined as follows:
\begin{equation}
\mathcal{L}_{CE}^i = -\sum_{c=1}^{3}\sum_{h=1}^{H}\sum_{w=1}^{W}{G(c,h,w)}log\left({P_s^{i}(c,h,w)}\right),
\label{equa:ce_loss}
\end{equation}
where $c$ represents the number of classes in the trimap.

Following~\cite{gfm}, for the matting decoder, we adopt alpha loss $\mathcal{L}_{\alpha}^M$ and Laplacian loss $\mathcal{L}_{lap}^M$ calculated only on the transition region. For the segmentation decoder, we adopt a cross-entropy loss $\mathcal{L}_{CE}^S$ on its final output. For the final output, we adopt alpha loss $\mathcal{L}_{\alpha}$, Laplacian loss $\mathcal{L}_{lap}$, and composition loss $\mathcal{L}_{comp}$ calculated on the whole image. The final training objective is a combination of all the aforementioned losses, $i.e.$,
\begin{equation}
\begin{aligned}
\mathcal{L} &= \lambda_m\left({\mathcal{L}_{\alpha}^M+\mathcal{L}_{lap}^M}\right)+
\lambda_{s}\left(\sum_{i=1}^{3}{L^i_{CE}}+3\mathcal{L}_{CE}^S\right)\\
    &\quad + \lambda_{f}\left(2\mathcal{L}_{\alpha}+2\mathcal{L}_{lap}+\mathcal{L}_{comp}\right),
\end{aligned}
\label{equa:loss}
\end{equation}
where $\lambda_m=2$, $\lambda_s=1/6$, and $\lambda_f=1$ are loss weights.

\begin{table*}[htb]
\begin{center}
\resizebox{\linewidth}{!}{
\begin{tabular}{c|cccccccc|cccccccc}
\hline
 & \multicolumn{8}{|c}{P3M-500-P} & \multicolumn{8}{|c}{P3M-500-NP} \\
\hline
Method & SAD & MSE & MAD & SAD-T& MSE-T& MAD-T& GRAD & CONN & SAD & MSE & MAD &SAD-T& MSE-T& MAD-T& GRAD & CONN\\
\hline
LF  & 42.95&0.0191 &0.0250 &12.43  &0.0421 &0.0824 &42.19  &18.80 &32.59&0.0131& 0.0188&14.53&0.0420&0.0825 & 31.93 & 19.50\\
HATT  & 25.99& 0.0054& 0.0152& 11.03 &0.0377 &0.0752 & 14.91 &25.29 &30.53&0.0072&0.0176&13.48&0.0403 & 0.0803 & 19.88 & 27.42\\
SHM  & 21.56 & 0.0100& 0.0125& 9.14 & 0.0255 & 0.0545 & 21.24 & 17.53&20.77& 0.0093&0.0122 & 9.14 & 0.0255 & 0.0545 & 20.30 & 17.09\\
GFM  & 13.20& 0.0050& 0.0080& 8.84 &0.0269 &0.0616 &12.58  & 17.75&15.50&0.0056&0.0091 & 10.16 & 0.0268 & 0.0620 & 14.82 & 18.03\\
\hline
DIM$\star$  & 4.89&0.0009 &0.0029 &4.89  &0.0115 &0.0342 &4.48  & 9.68& 5.32&0.0009& 0.0031&5.32 & 0.0094 & 0.0324 & 4.70 & 7.70\\
 \hline
P3M-Net  & 8.73& 0.0026& 0.0051&6.89  &0.0193 &0.0478 &8.22  &13.88 &11.23&0.0035&0.0065 &7.65 & 0.0173 & 0.0466 & 10.35 & 12.51\\ 
\hline
\end{tabular}}
\end{center}
\caption{Results of P3M-Net and other methods on P3M-500-P and P3M-500-NP. DIM$\star$ uses ground truth trimap. }
\label{tab:experiment}
\end{table*}

\begin{figure*}[!t]
\centering
\captionsetup[subfloat]{labelformat=empty,justification=centering}

\subfloat[]{\includegraphics[width=.125\linewidth]{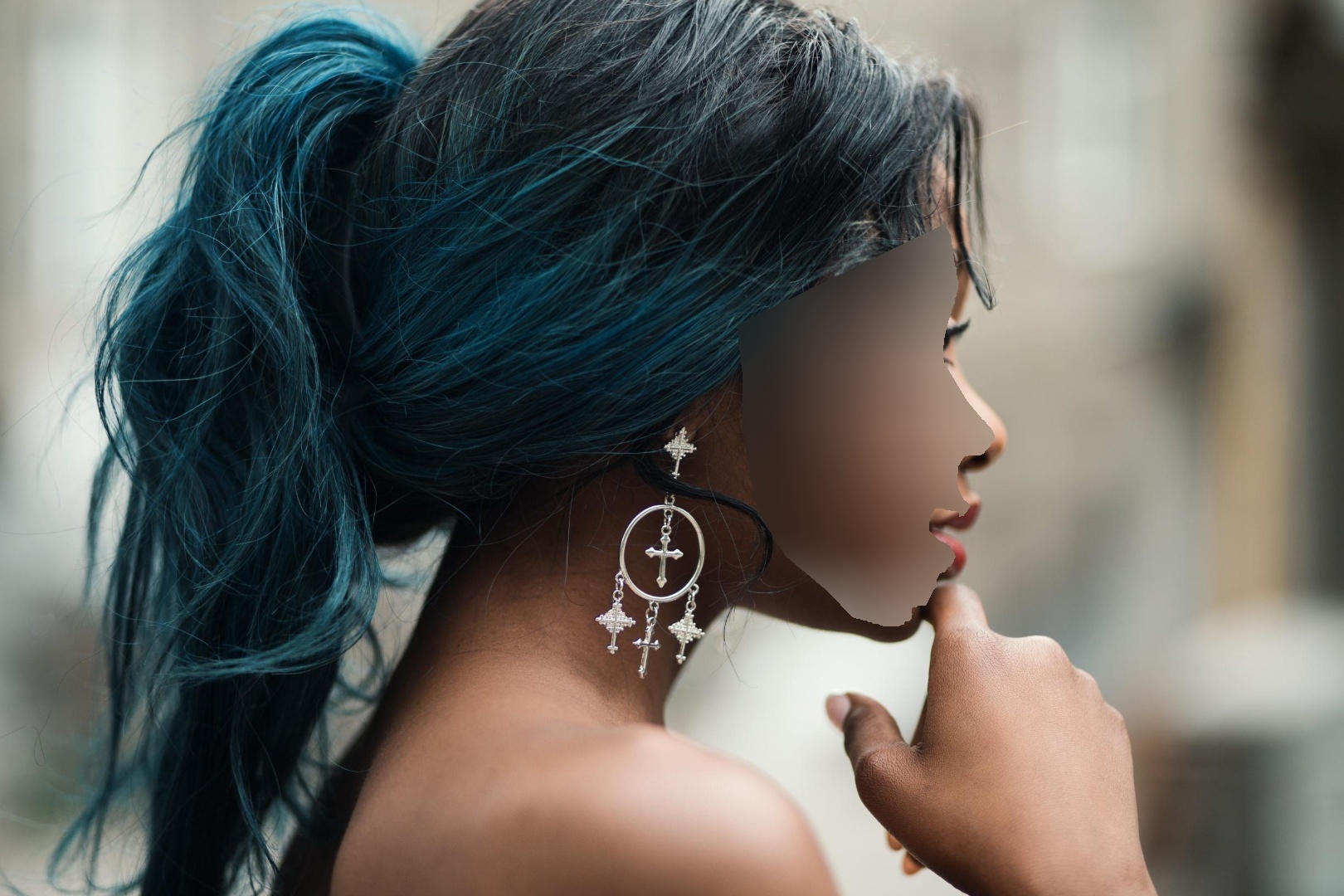}}
\subfloat[]{\includegraphics[width=.125\linewidth]{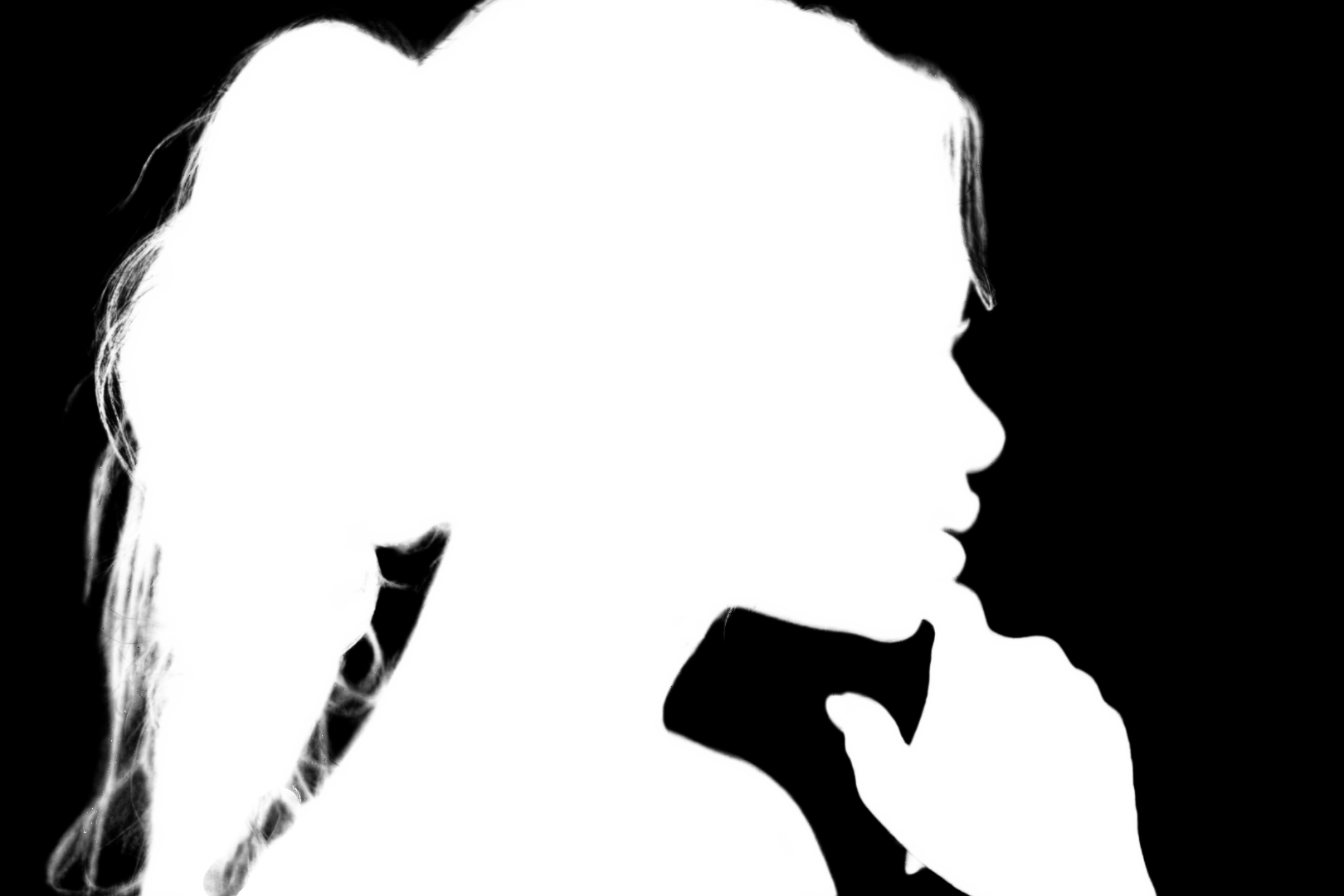}}
\subfloat[]{\includegraphics[width=.125\linewidth]{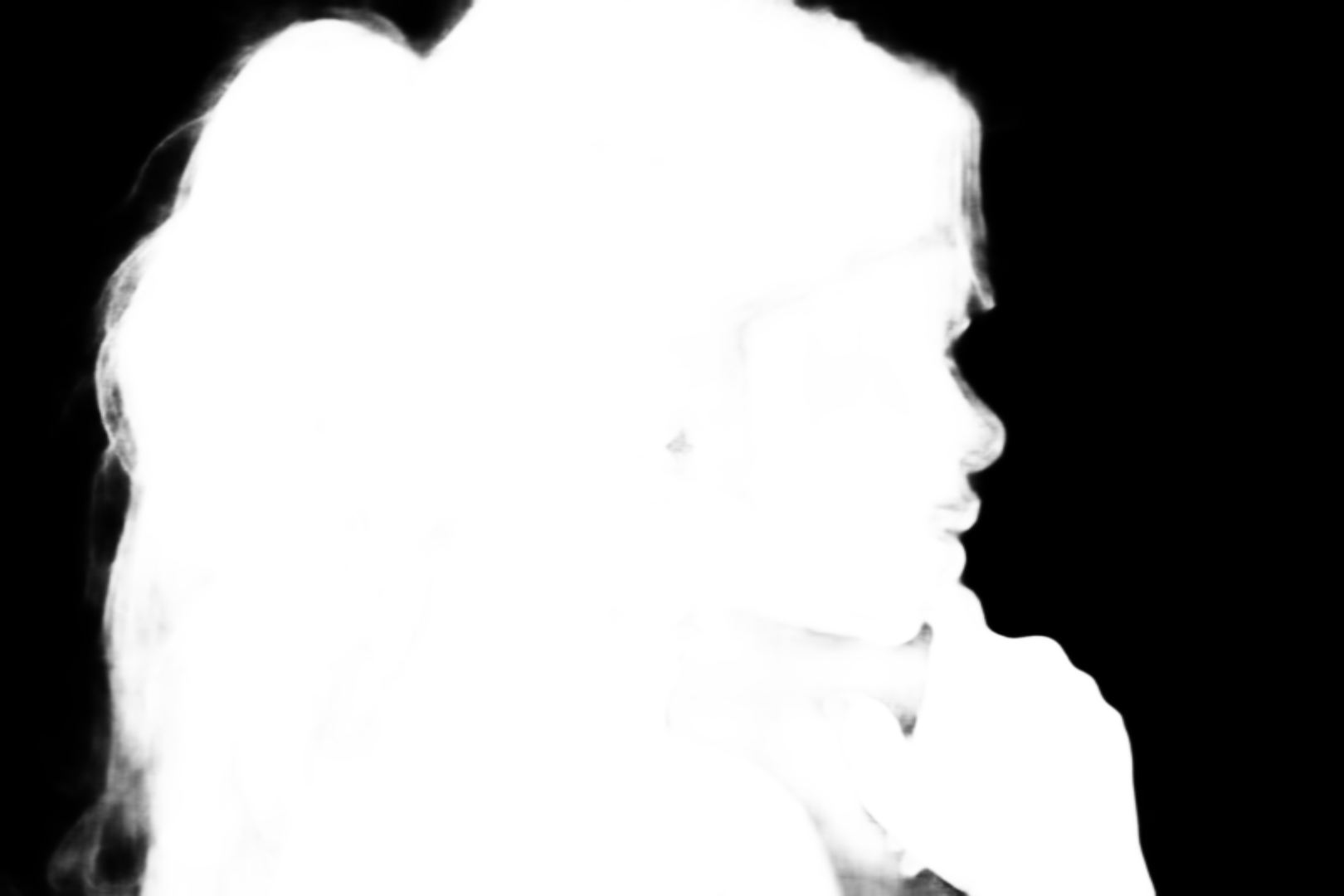}}
\subfloat[]{\includegraphics[width=.125\linewidth]{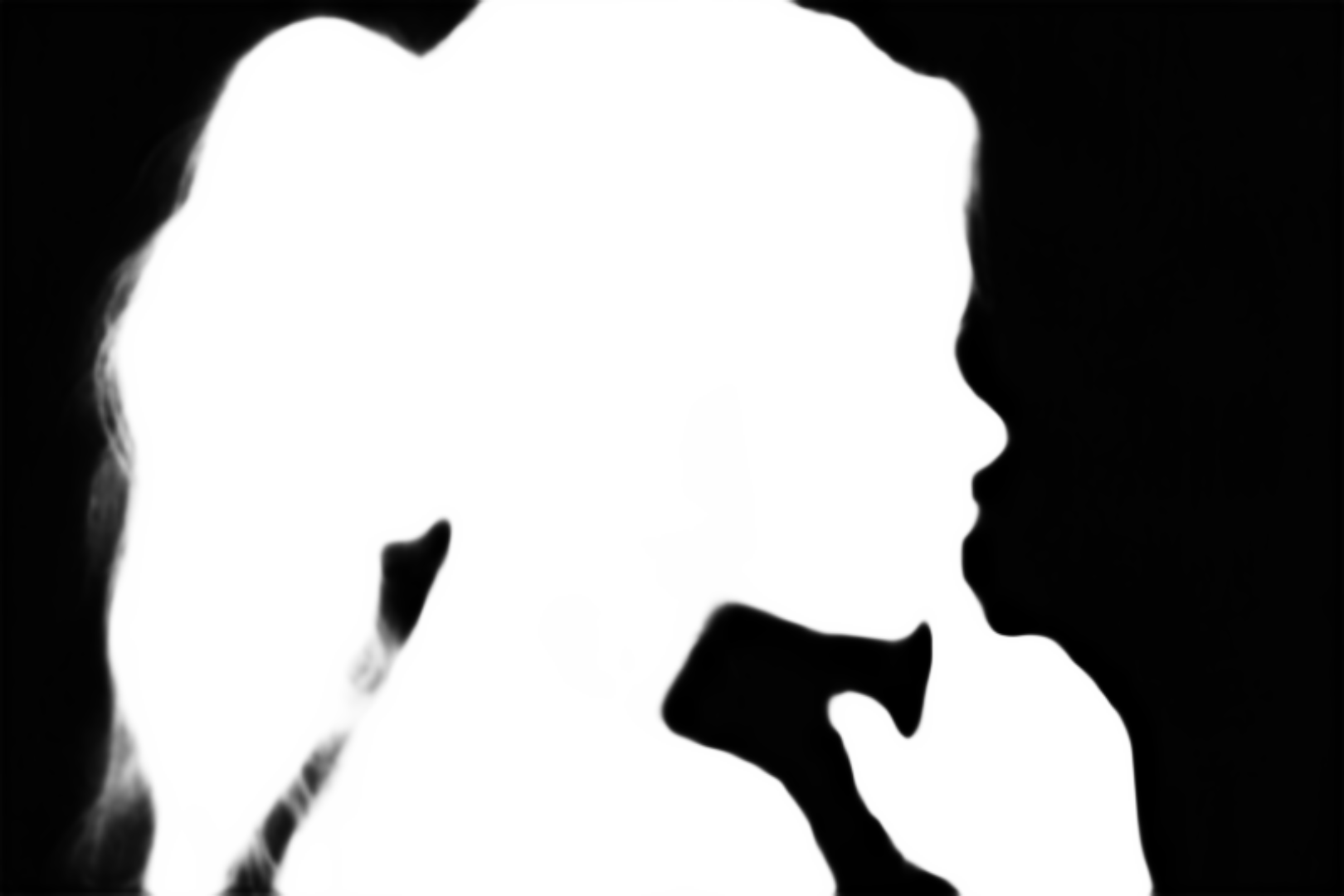}}
\subfloat[]{\includegraphics[width=.125\linewidth]{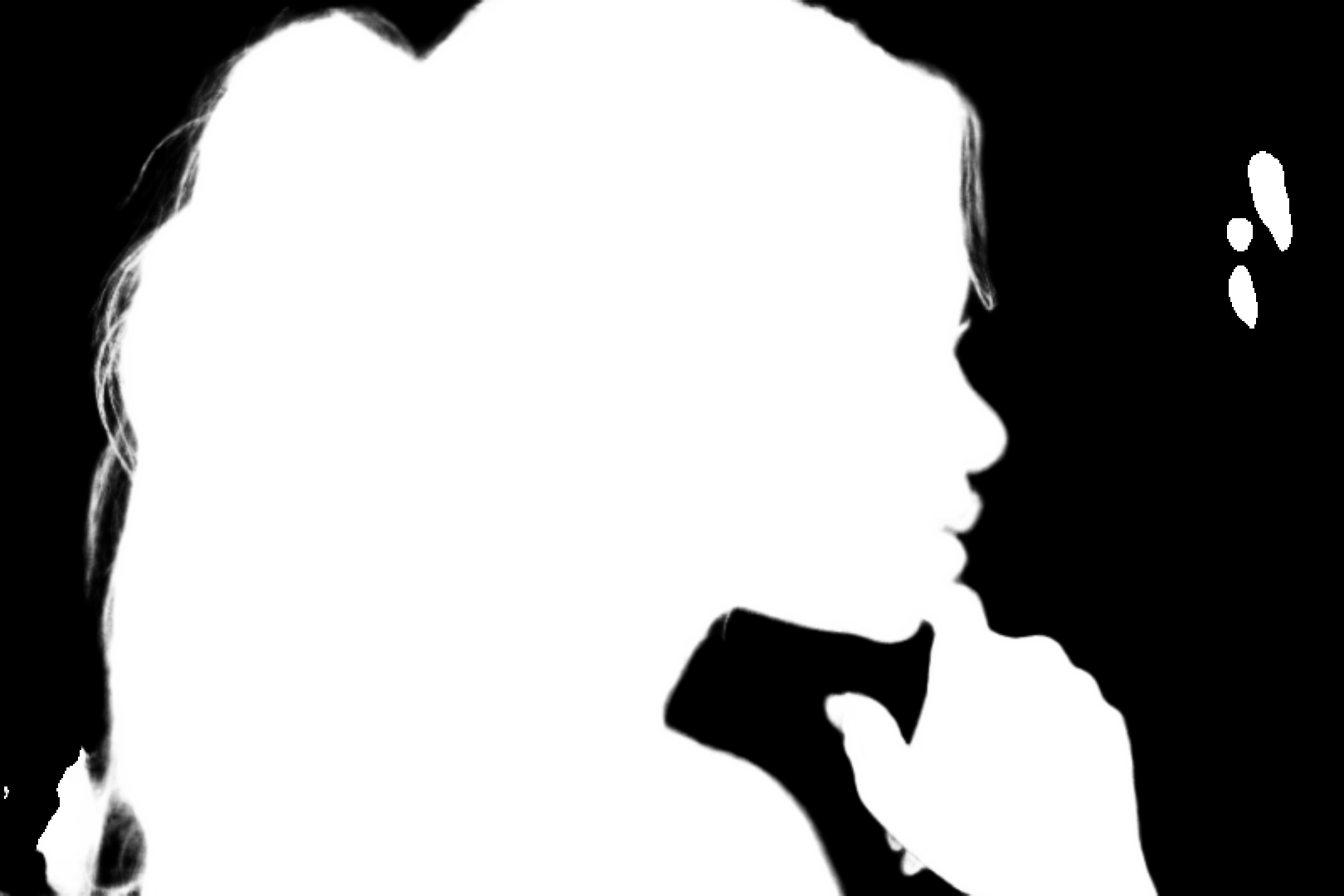}}
\subfloat[]{\includegraphics[width=.125\linewidth]{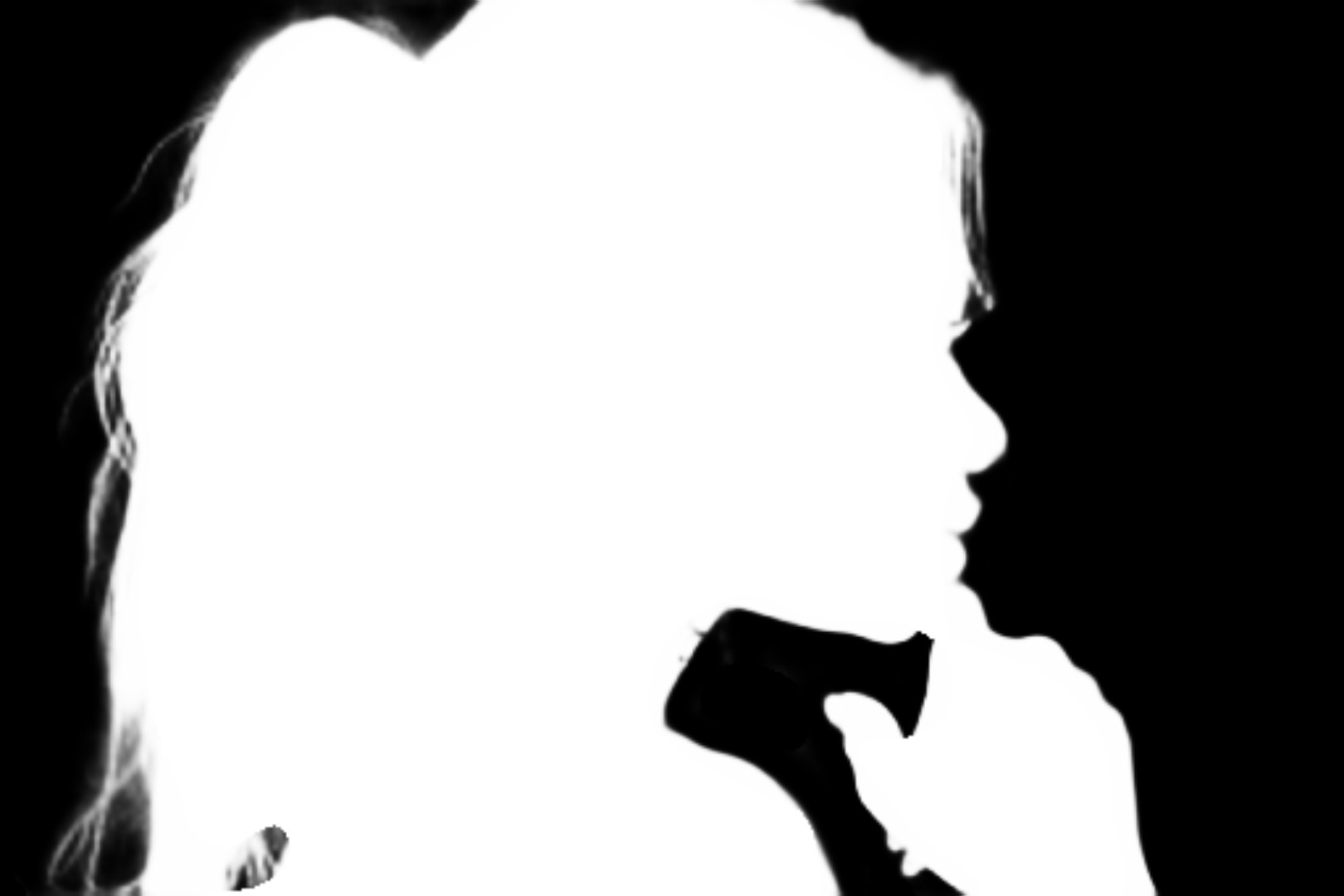}}
\subfloat[]{\includegraphics[width=.125\linewidth]{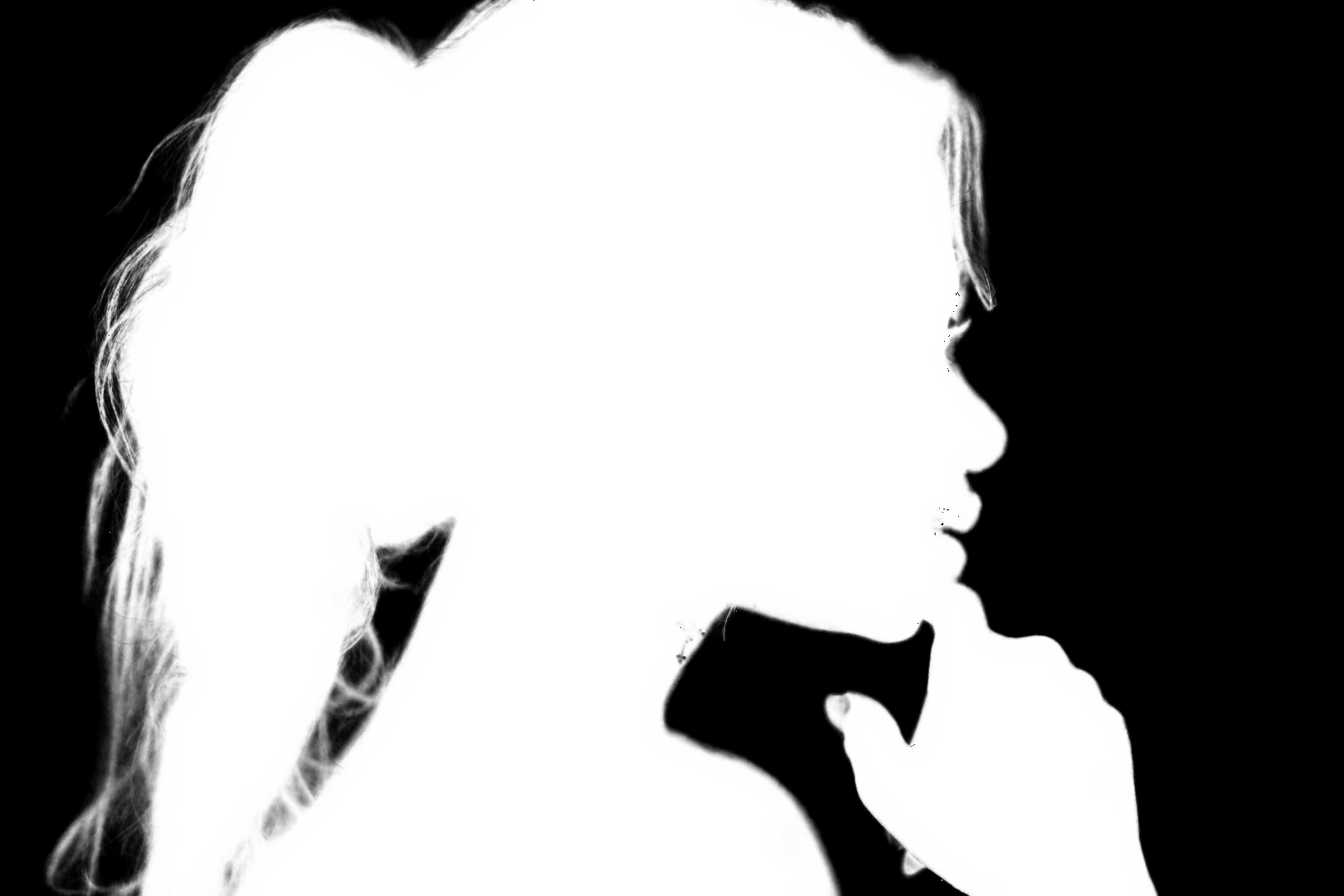}}
\subfloat[]{\includegraphics[width=.125\linewidth]{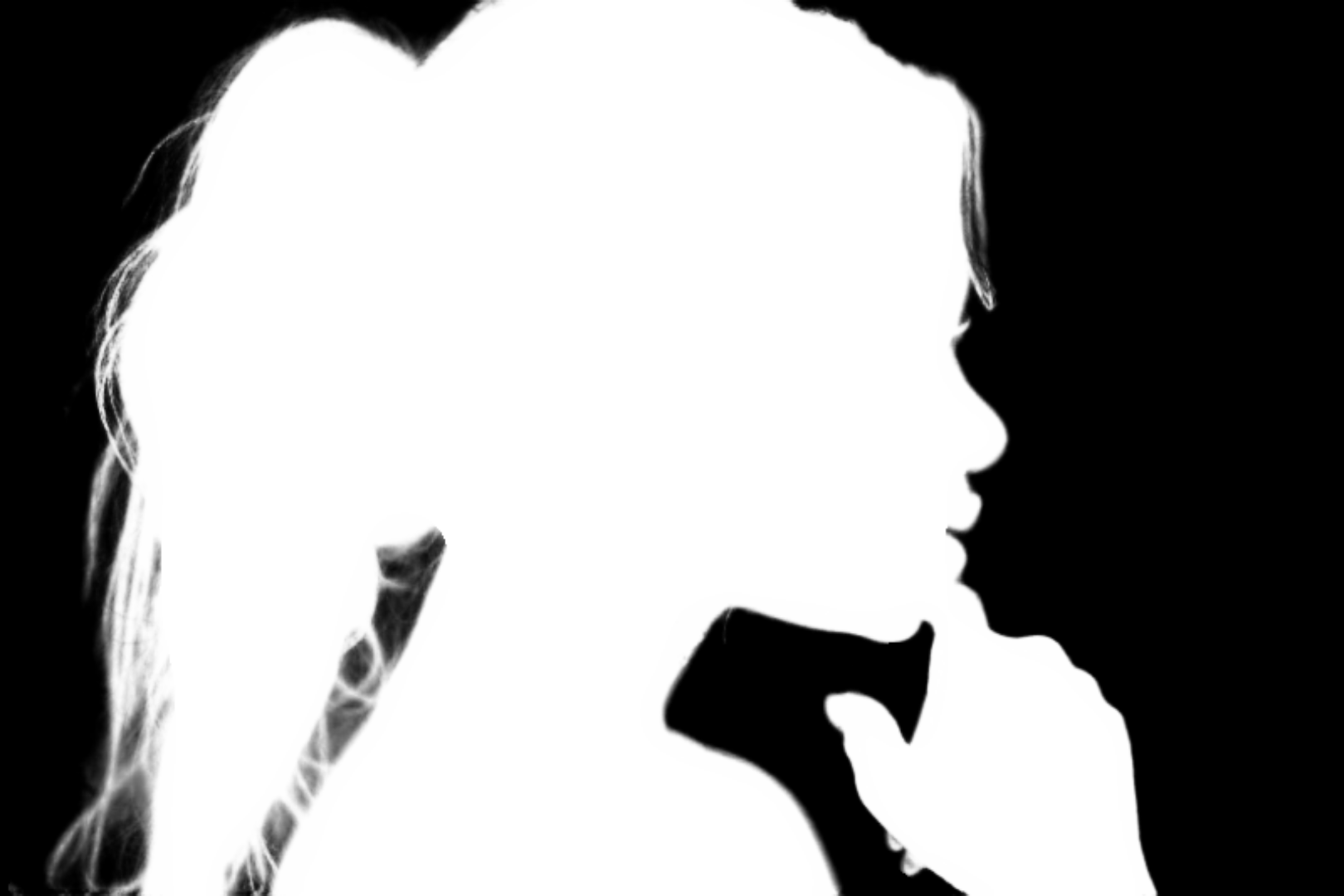}}\\[-6ex]

\subfloat[]{\includegraphics[width=.125\linewidth]{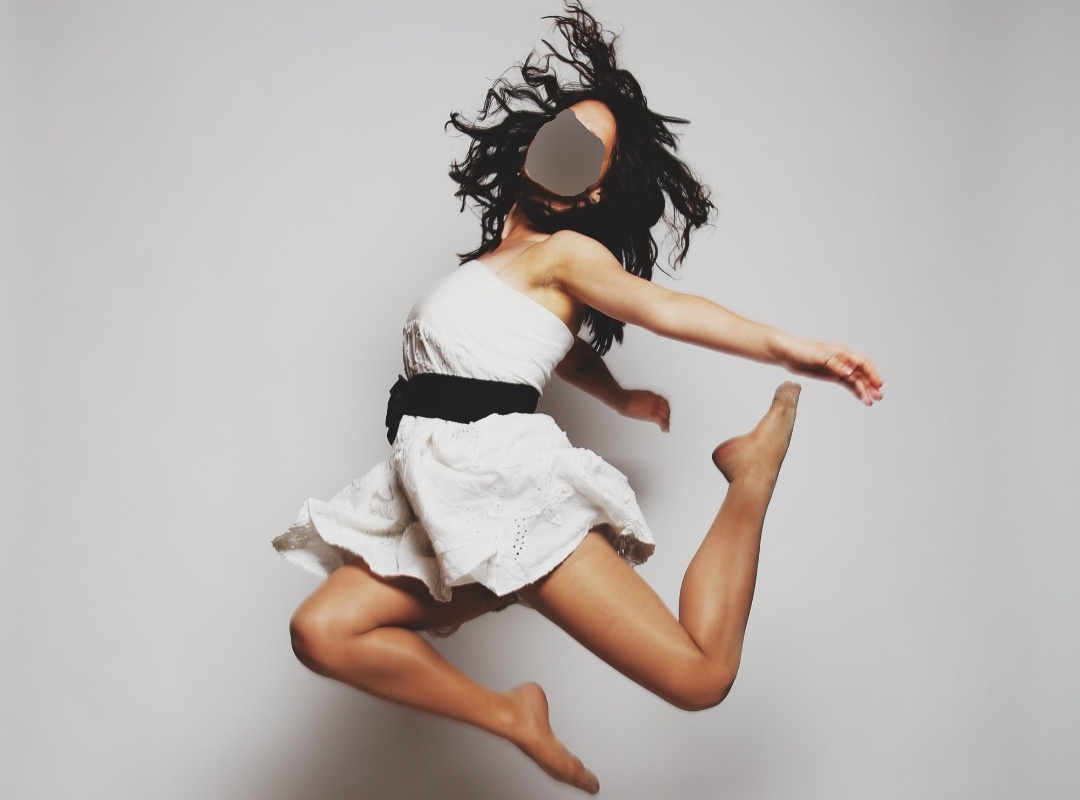}}
\subfloat[]{\includegraphics[width=.125\linewidth]{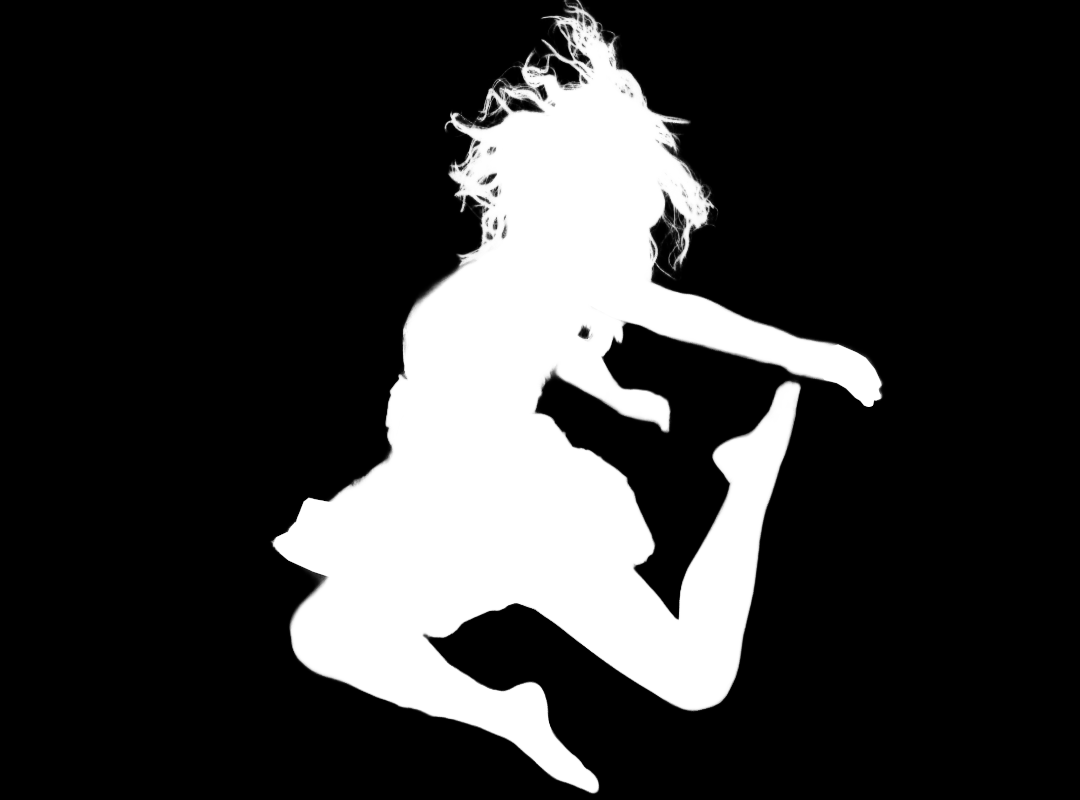}}
\subfloat[]{\includegraphics[width=.125\linewidth]{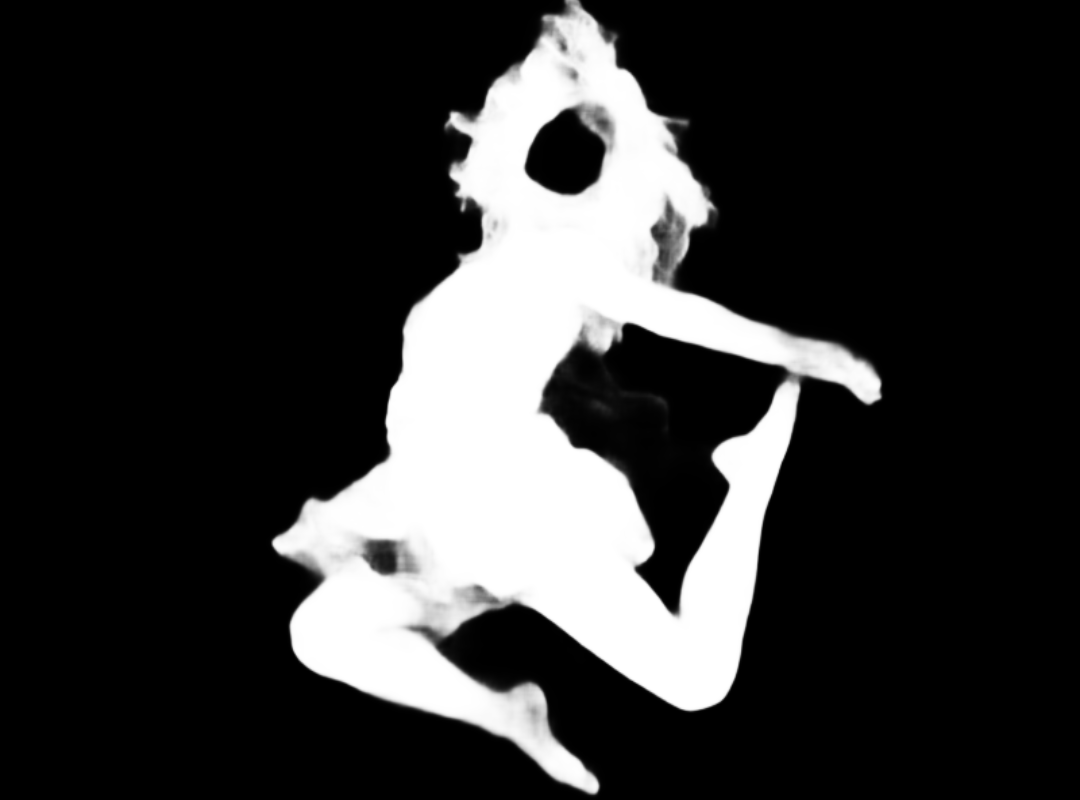}}
\subfloat[]{\includegraphics[width=.125\linewidth]{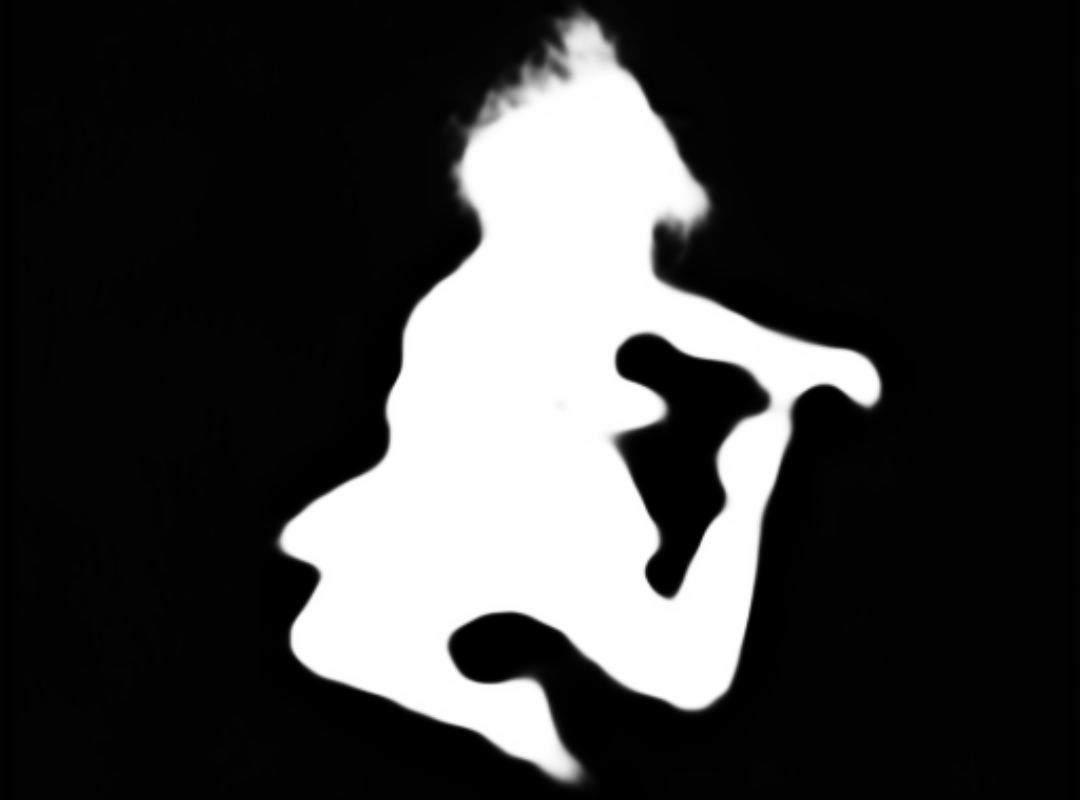}}
\subfloat[]{\includegraphics[width=.125\linewidth]{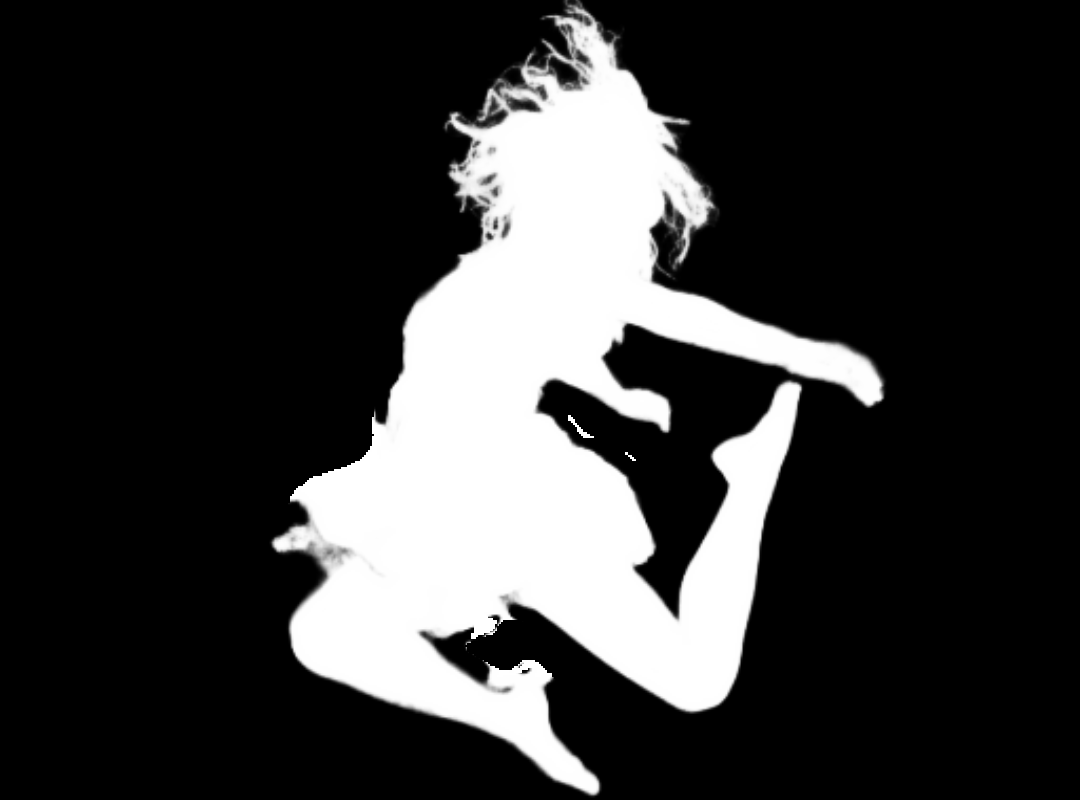}}
\subfloat[]{\includegraphics[width=.125\linewidth]{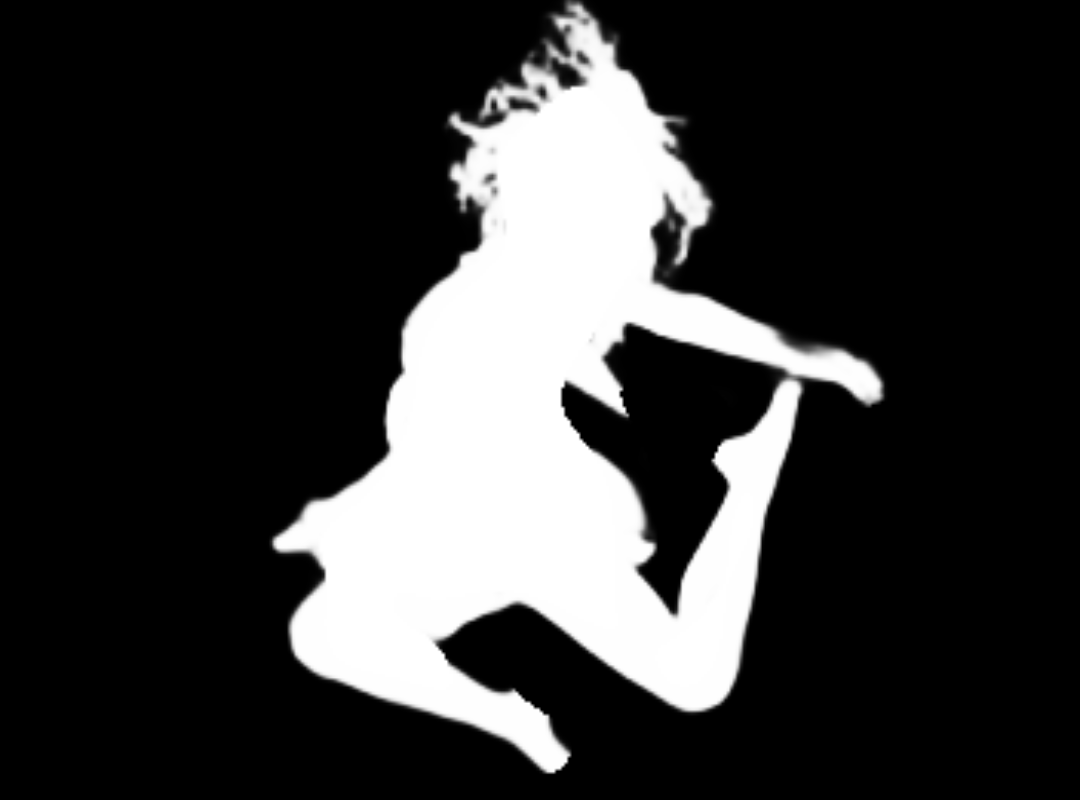}}
\subfloat[]{\includegraphics[width=.125\linewidth]{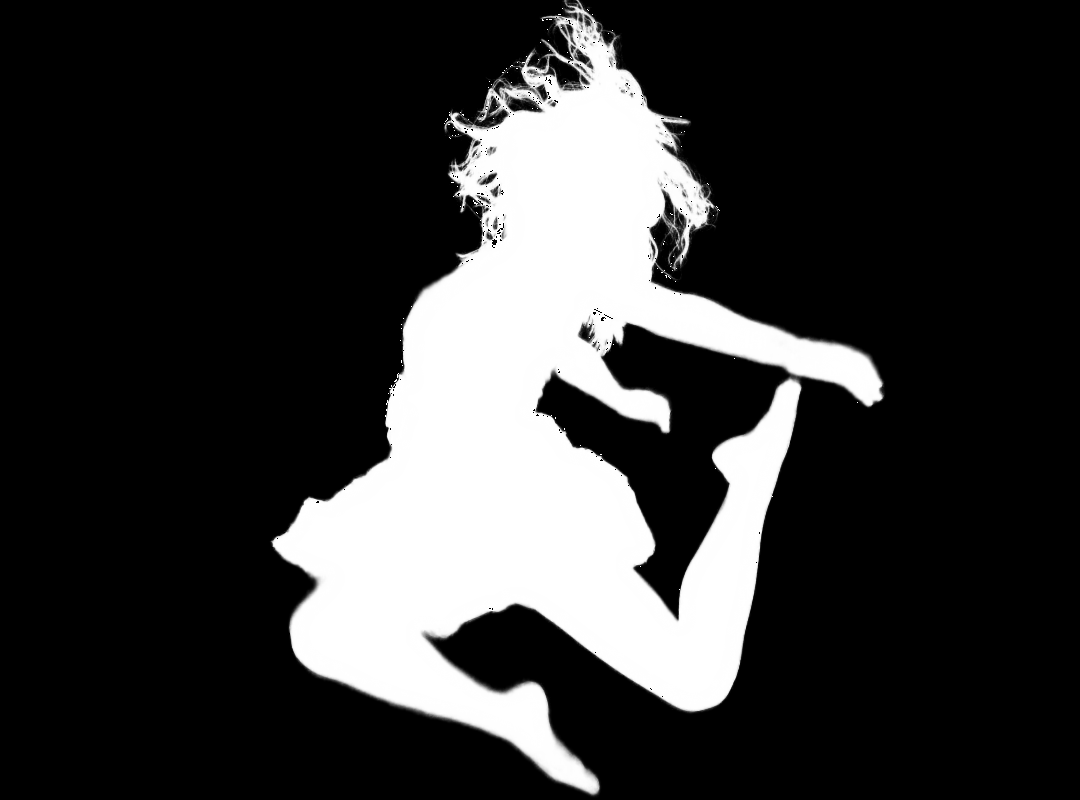}}
\subfloat[]{\includegraphics[width=.125\linewidth]{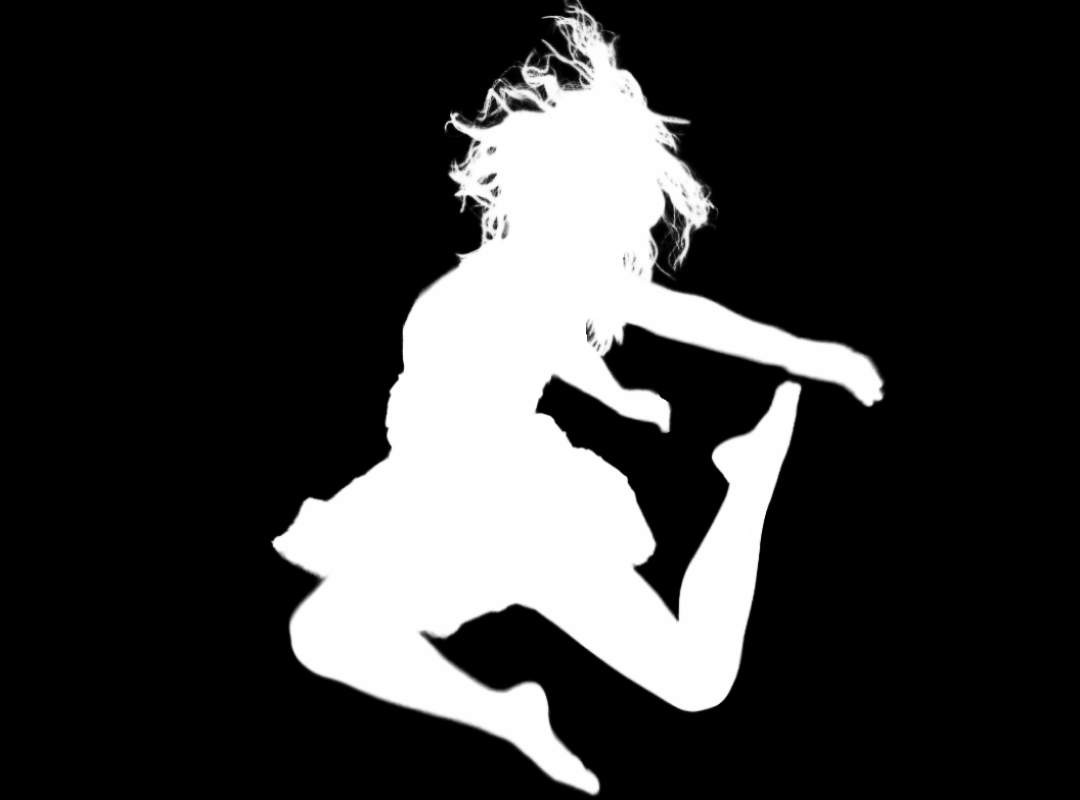}}\\[-6ex]

\subfloat[]{\includegraphics[width=.125\linewidth]{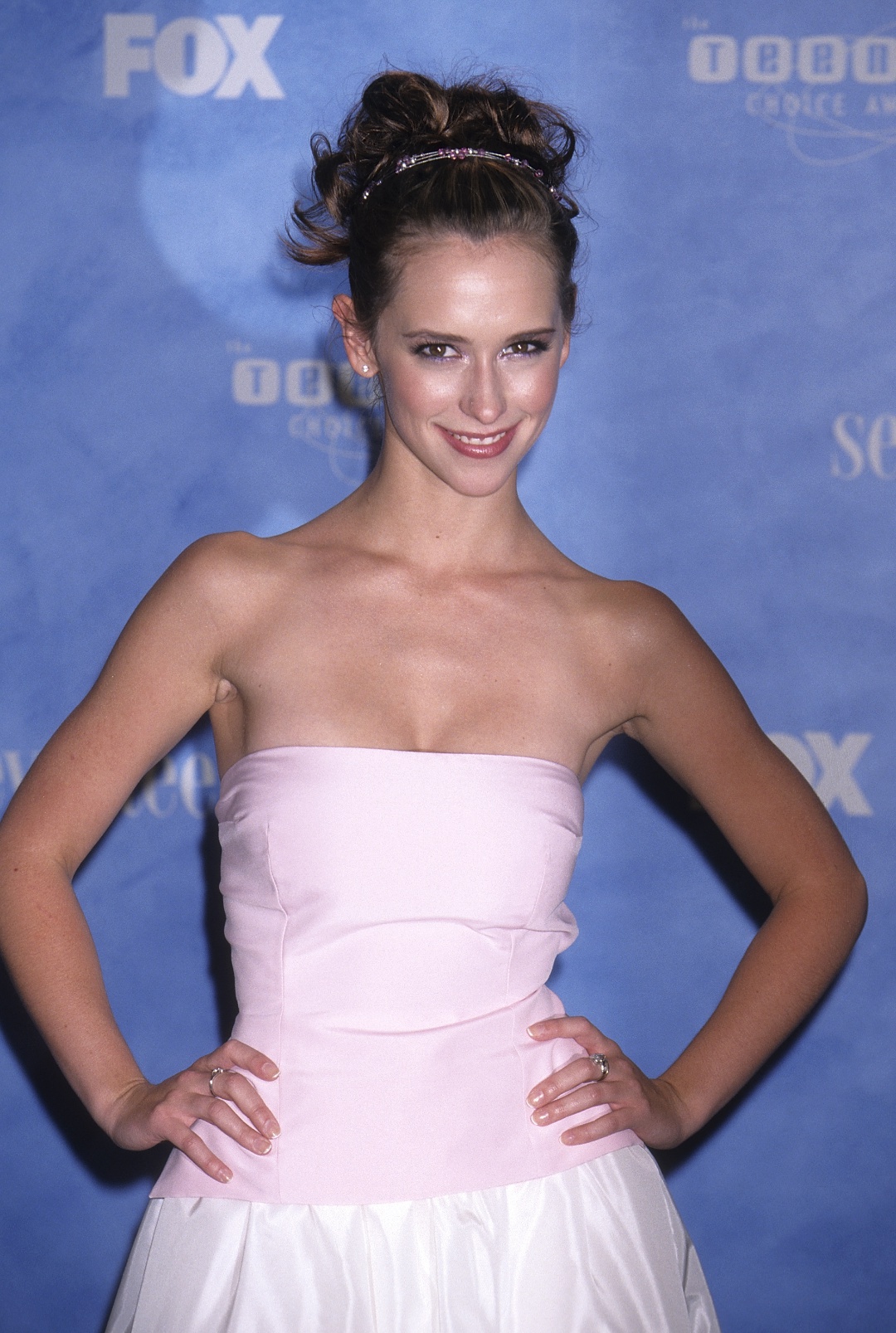}}
\subfloat[]{\includegraphics[width=.125\linewidth]{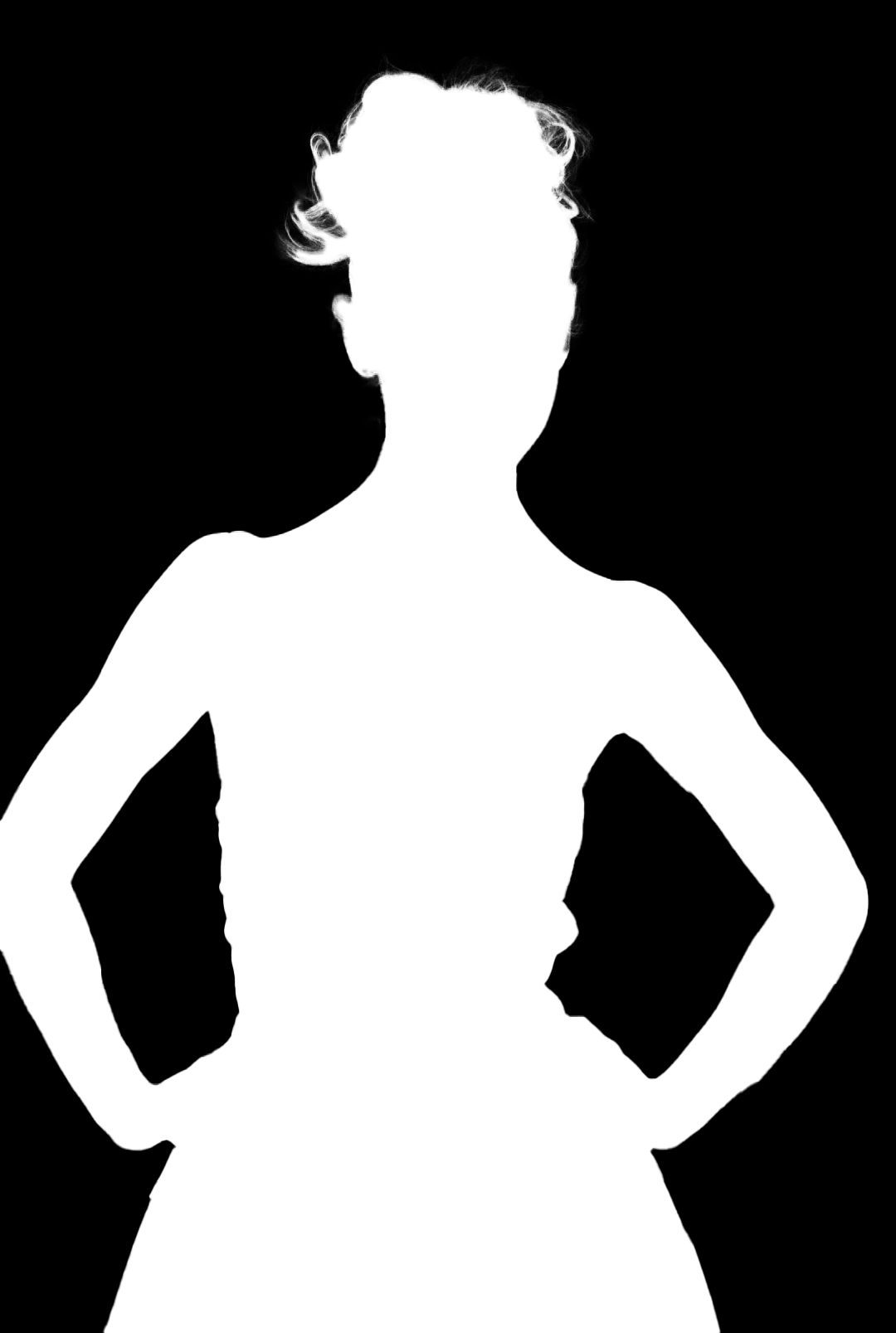}}
\subfloat[]{\includegraphics[width=.125\linewidth]{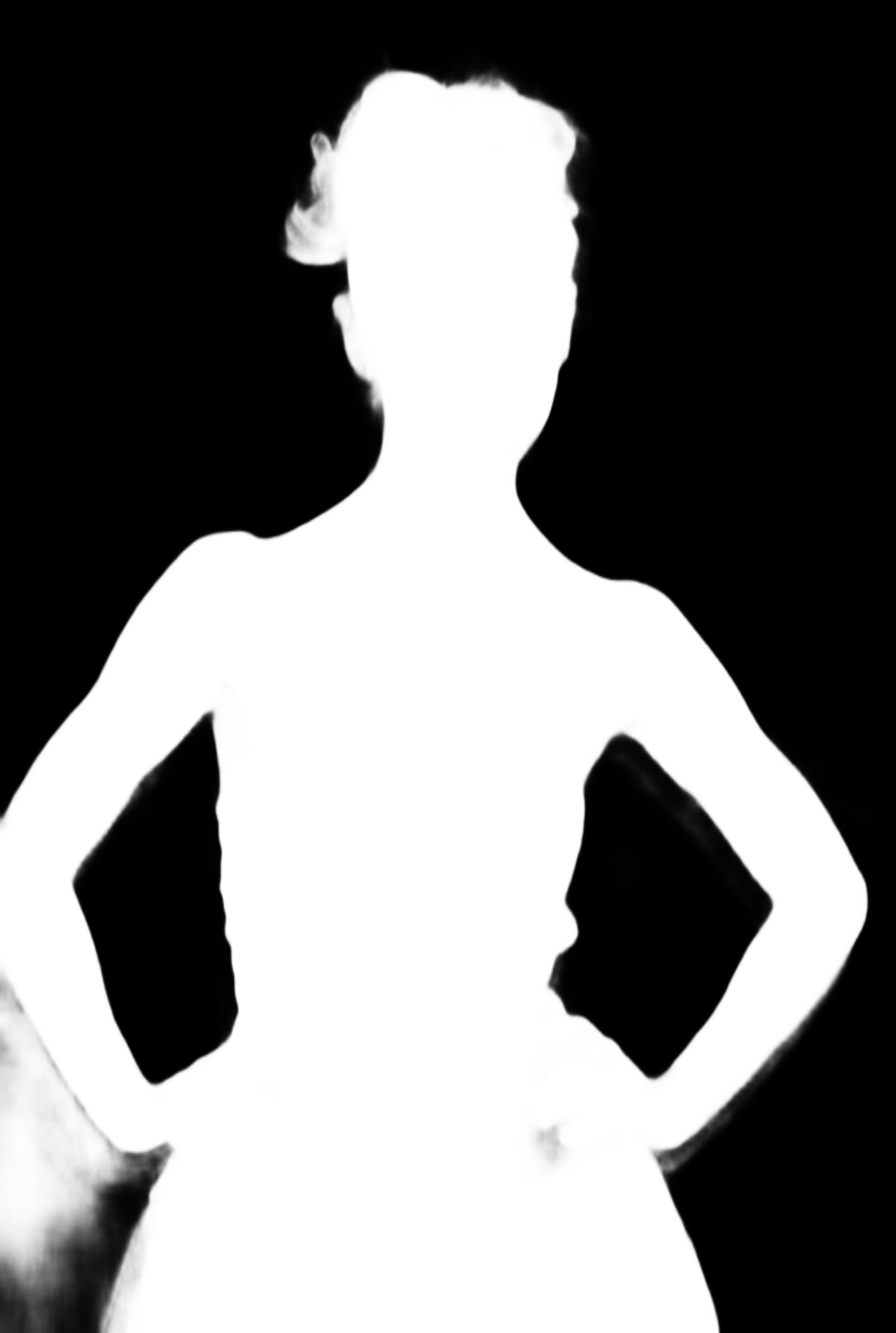}}
\subfloat[]{\includegraphics[width=.125\linewidth]{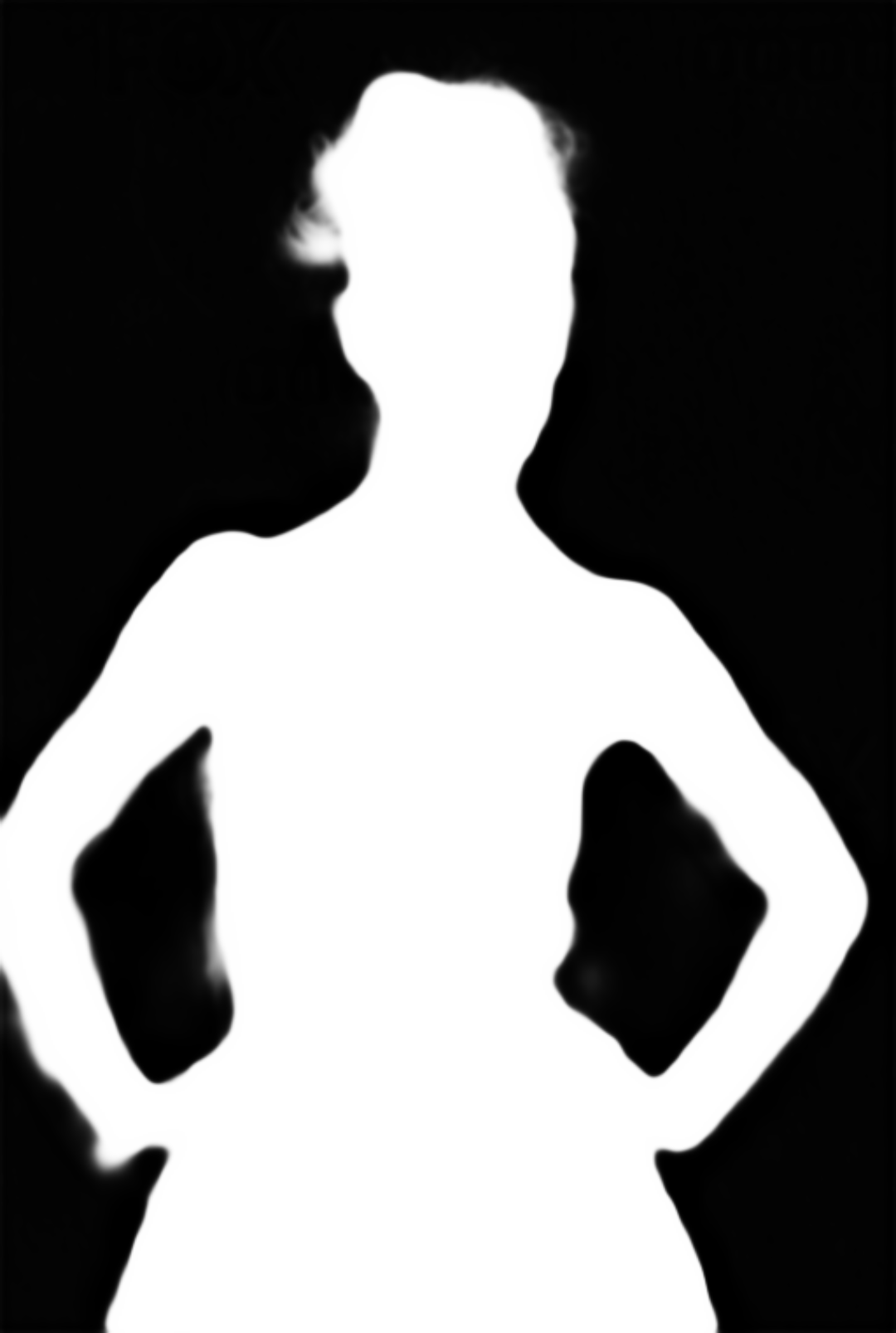}}
\subfloat[]{\includegraphics[width=.125\linewidth]{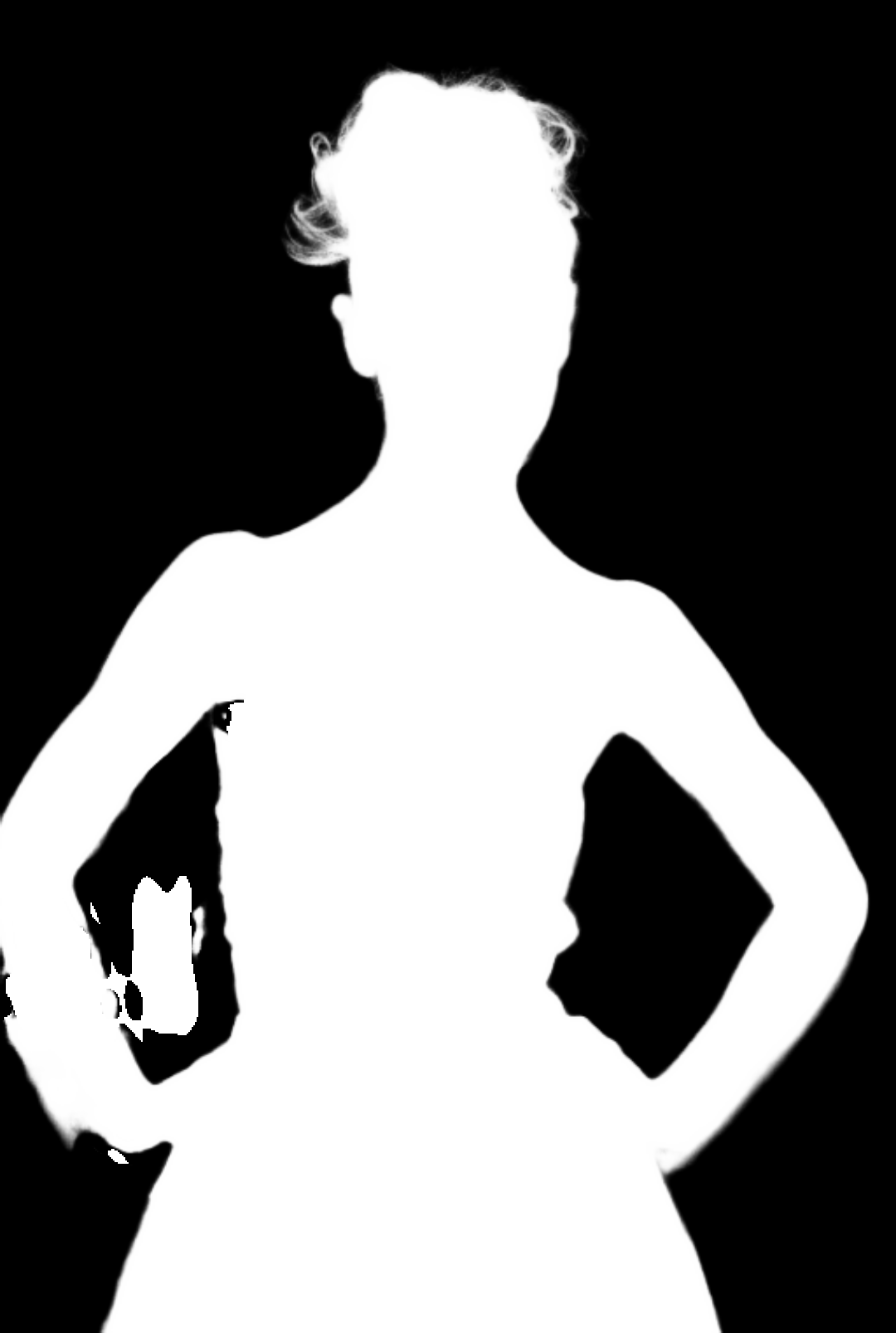}}
\subfloat[]{\includegraphics[width=.125\linewidth]{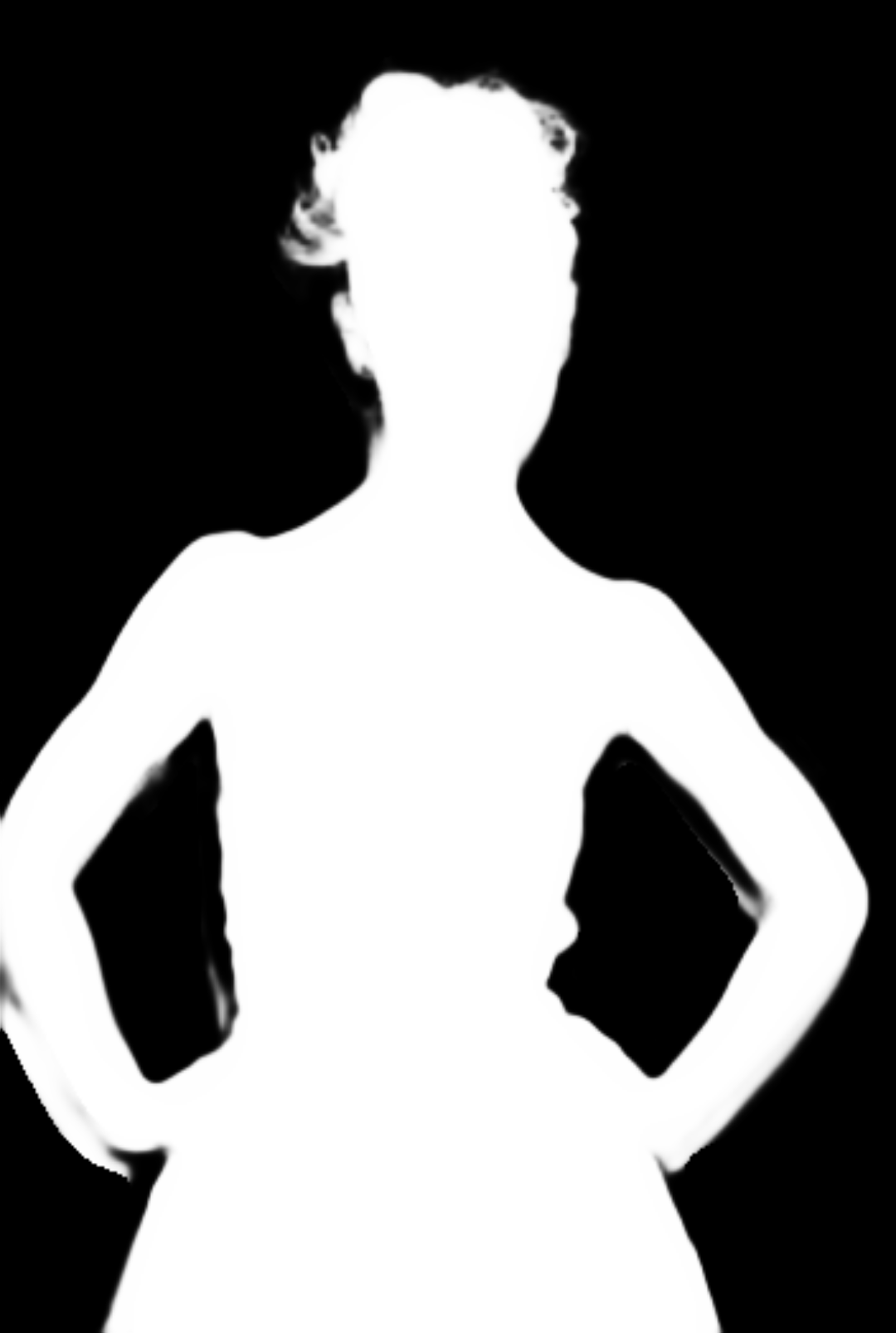}}
\subfloat[]{\includegraphics[width=.125\linewidth]{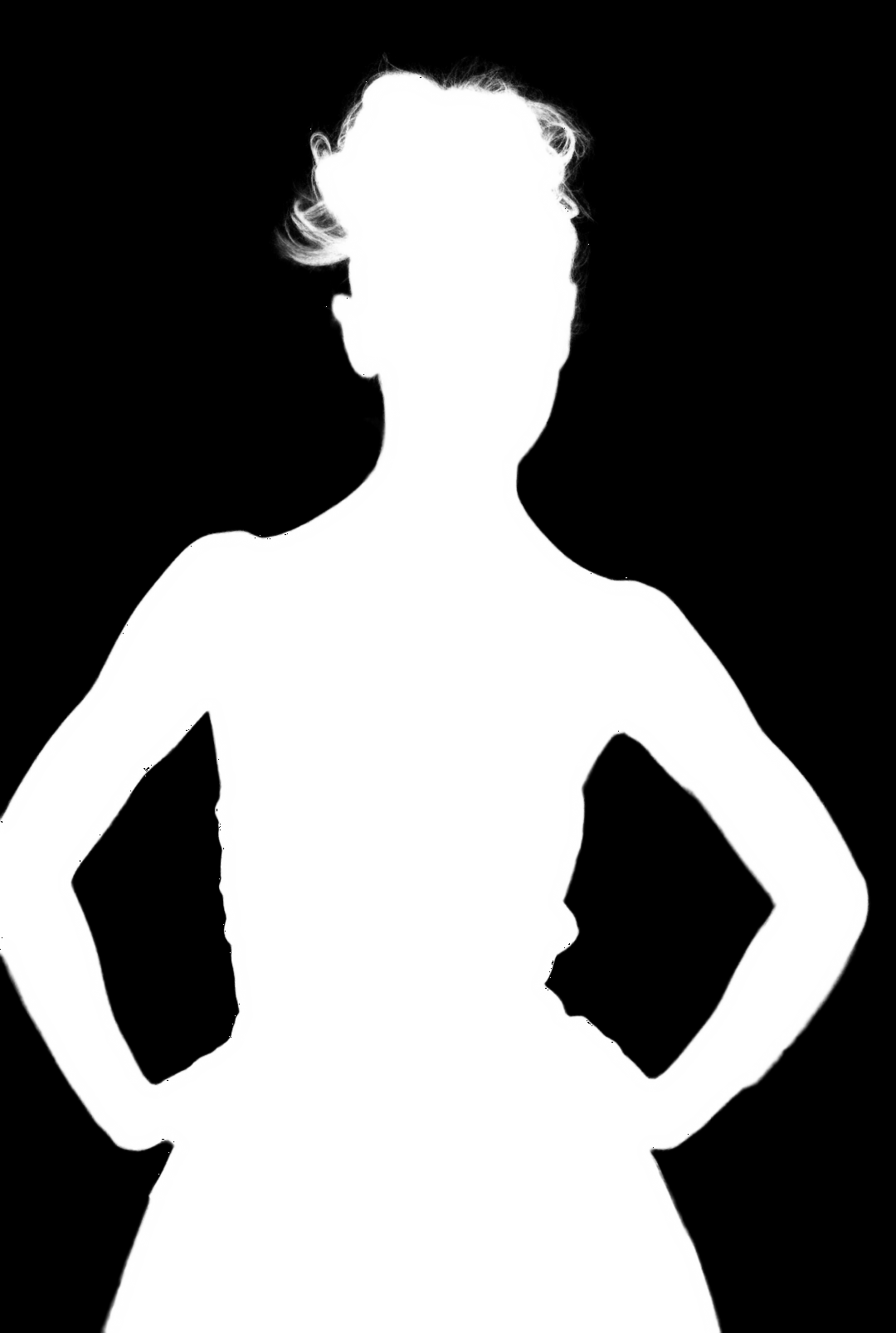}}
\subfloat[]{\includegraphics[width=.125\linewidth]{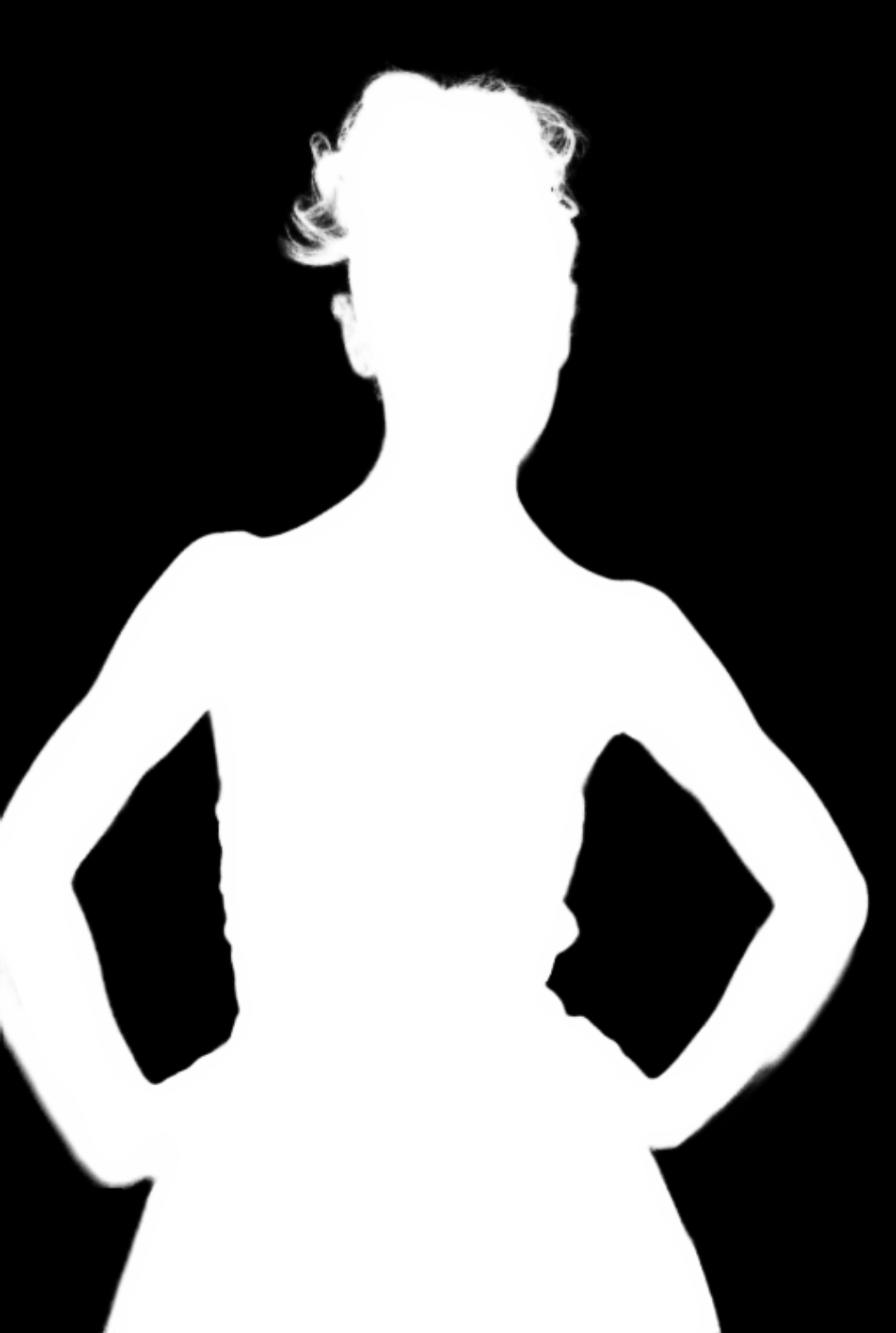}}\\[-6ex]

\subfloat[Image]{\includegraphics[width=.125\linewidth]{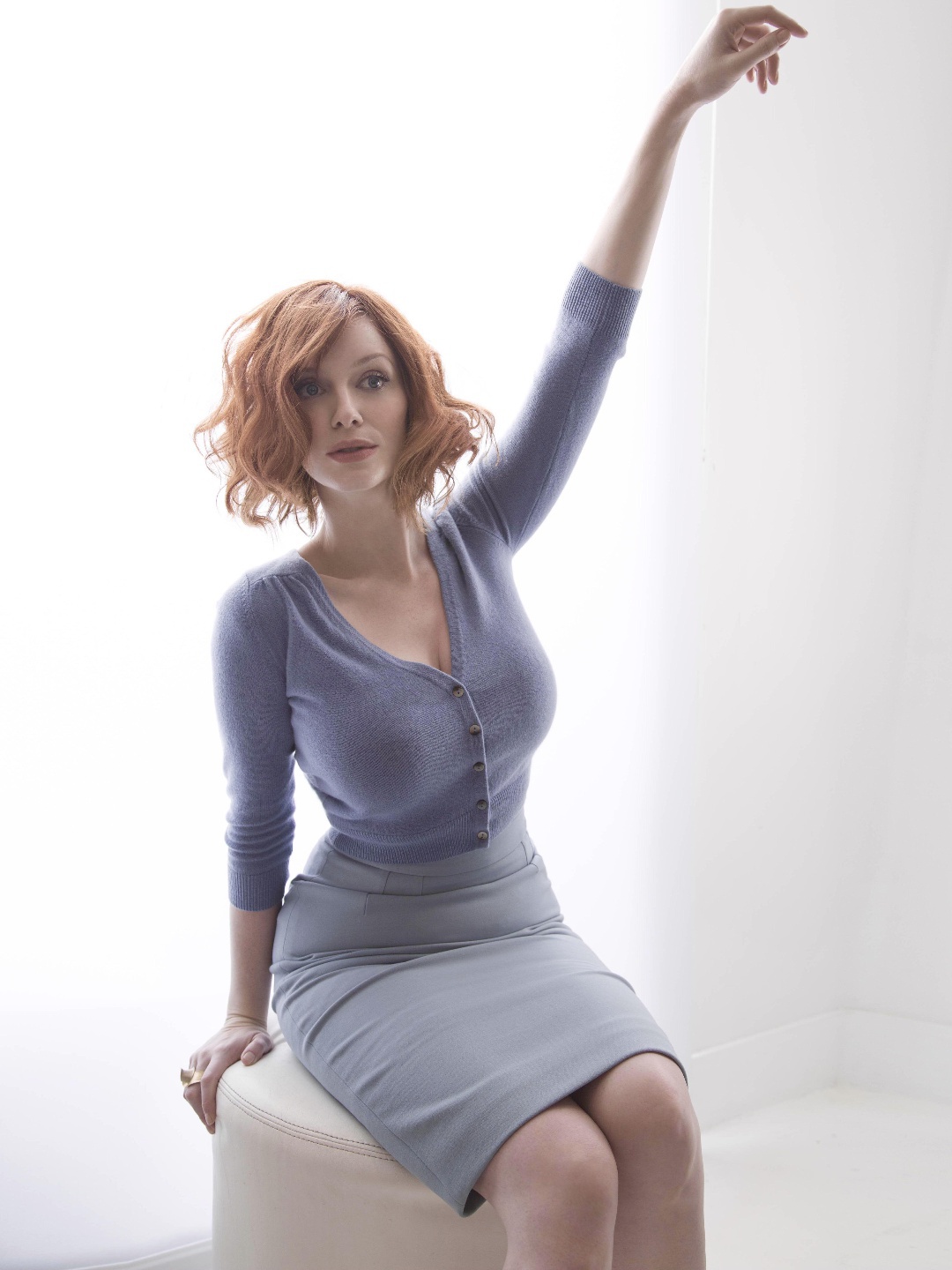}}
\subfloat[GT]{\includegraphics[width=.125\linewidth]{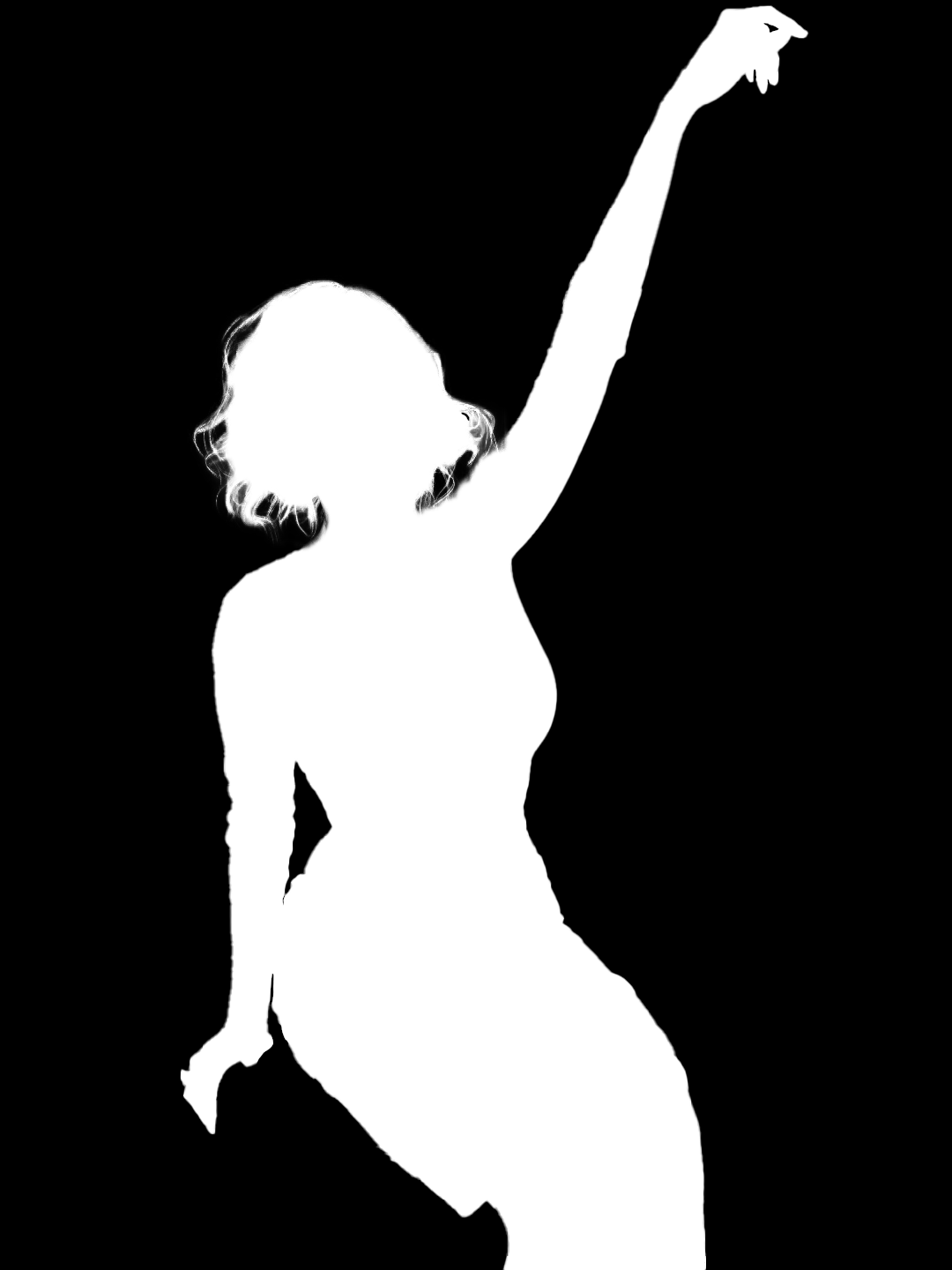}}
\subfloat[LF~\cite{lf}]{\includegraphics[width=.125\linewidth]{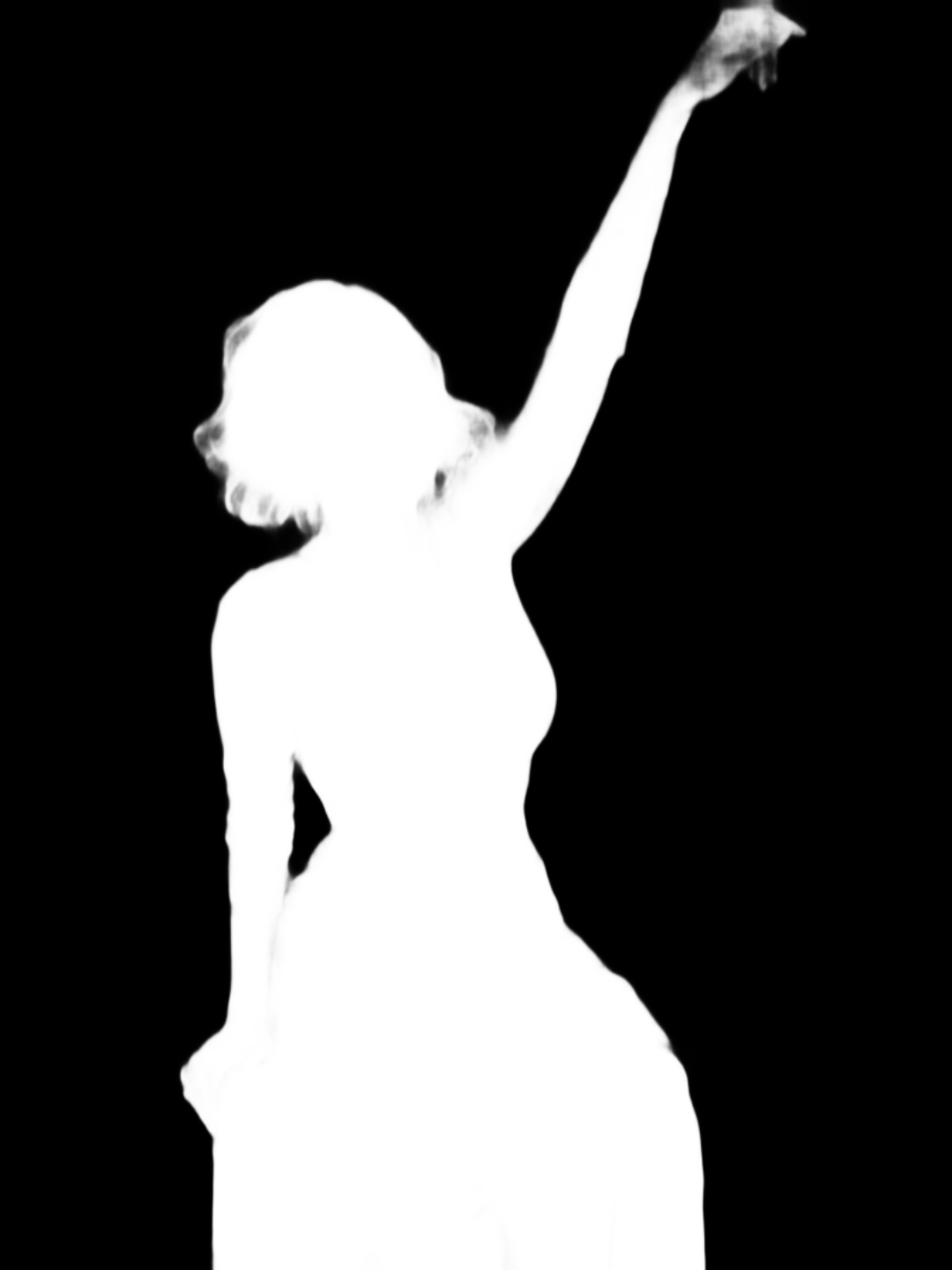}}
\subfloat[HATT~\cite{hatt}]{\includegraphics[width=.125\linewidth]{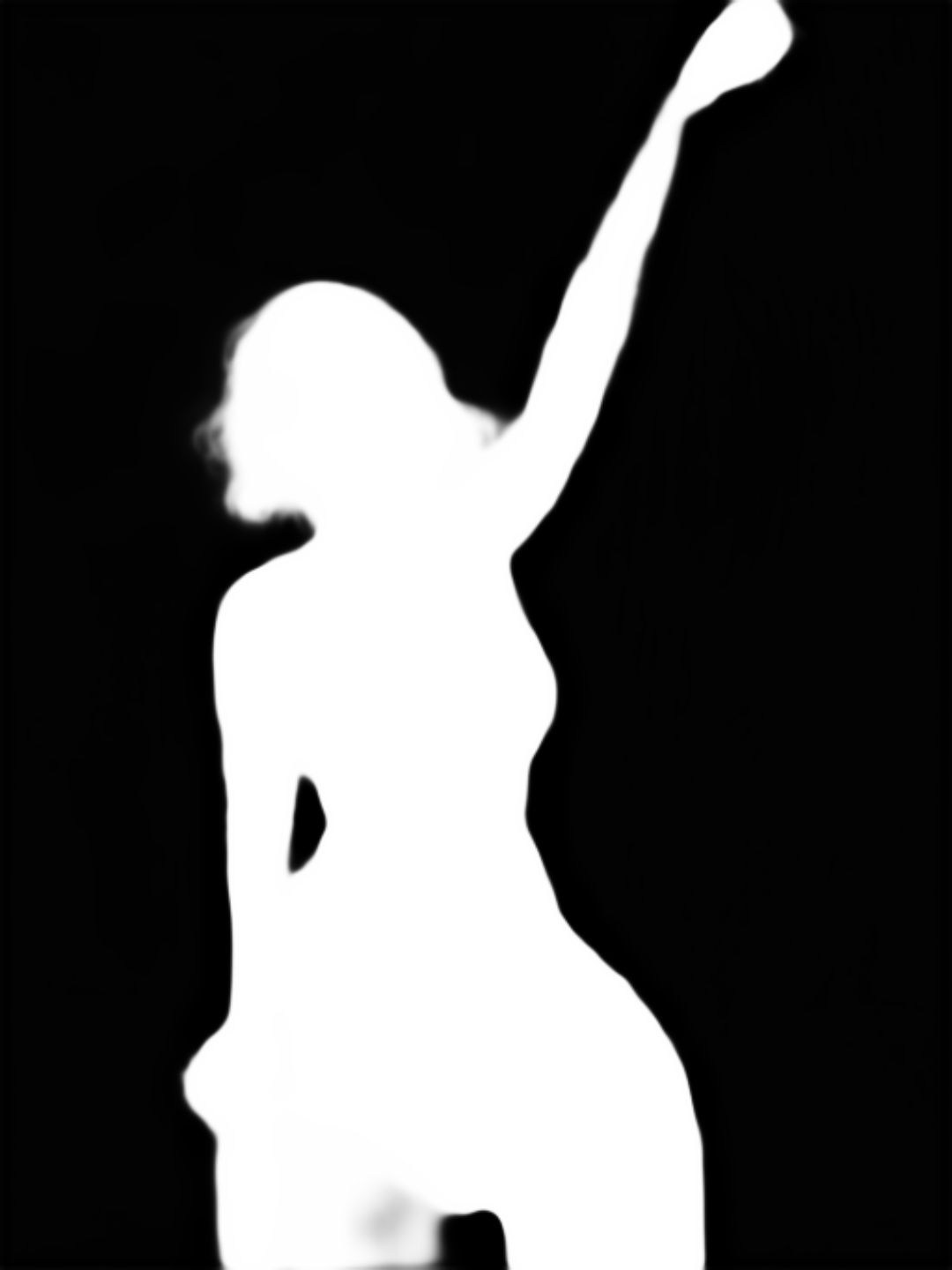}}
\subfloat[SHM~\cite{shm}]{\includegraphics[width=.125\linewidth]{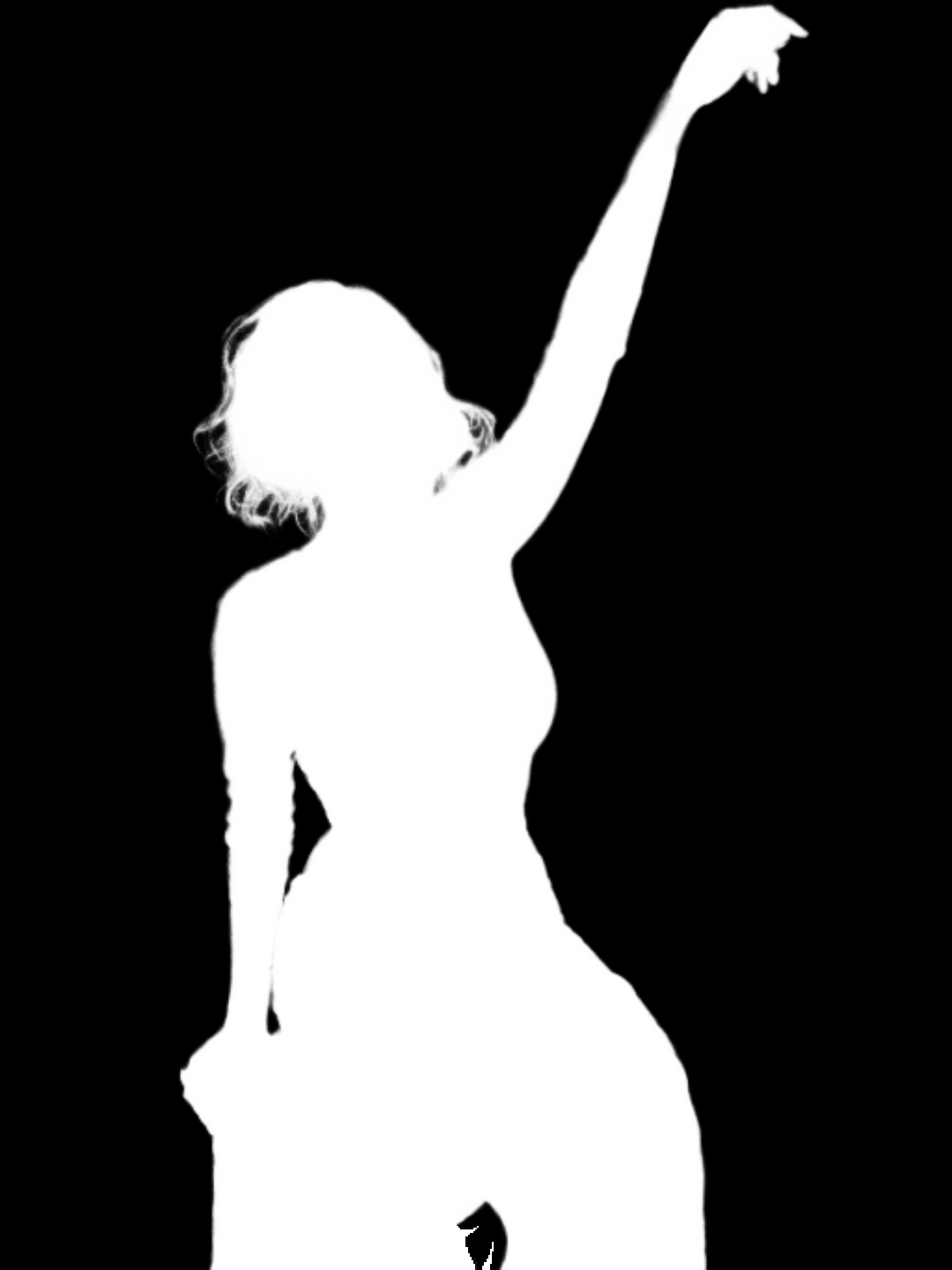}}
\subfloat[GFM~\cite{gfm}]{\includegraphics[width=.125\linewidth]{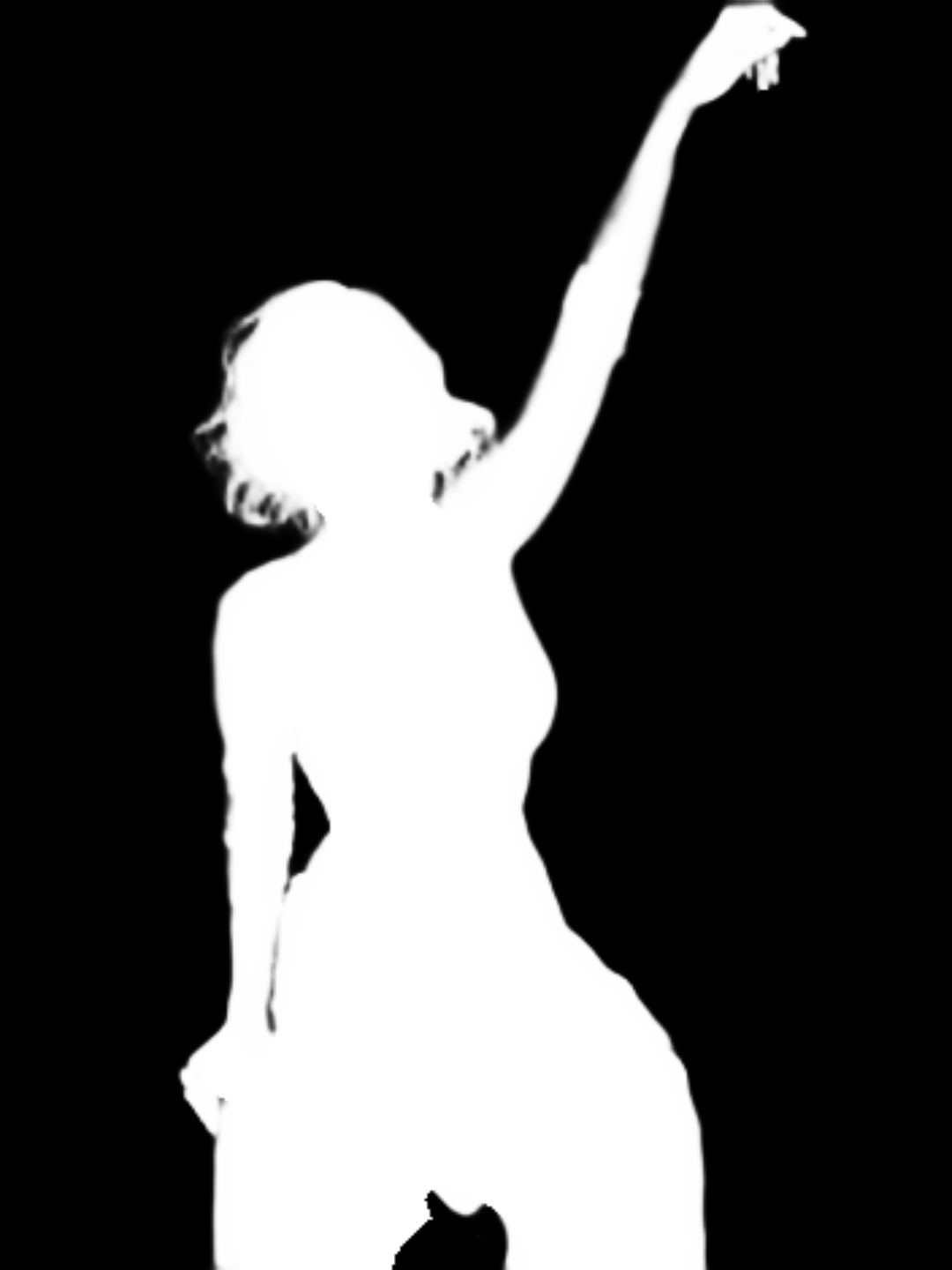}}
\subfloat[DIM~\cite{dim}+Trimap]{\includegraphics[width=.125\linewidth]{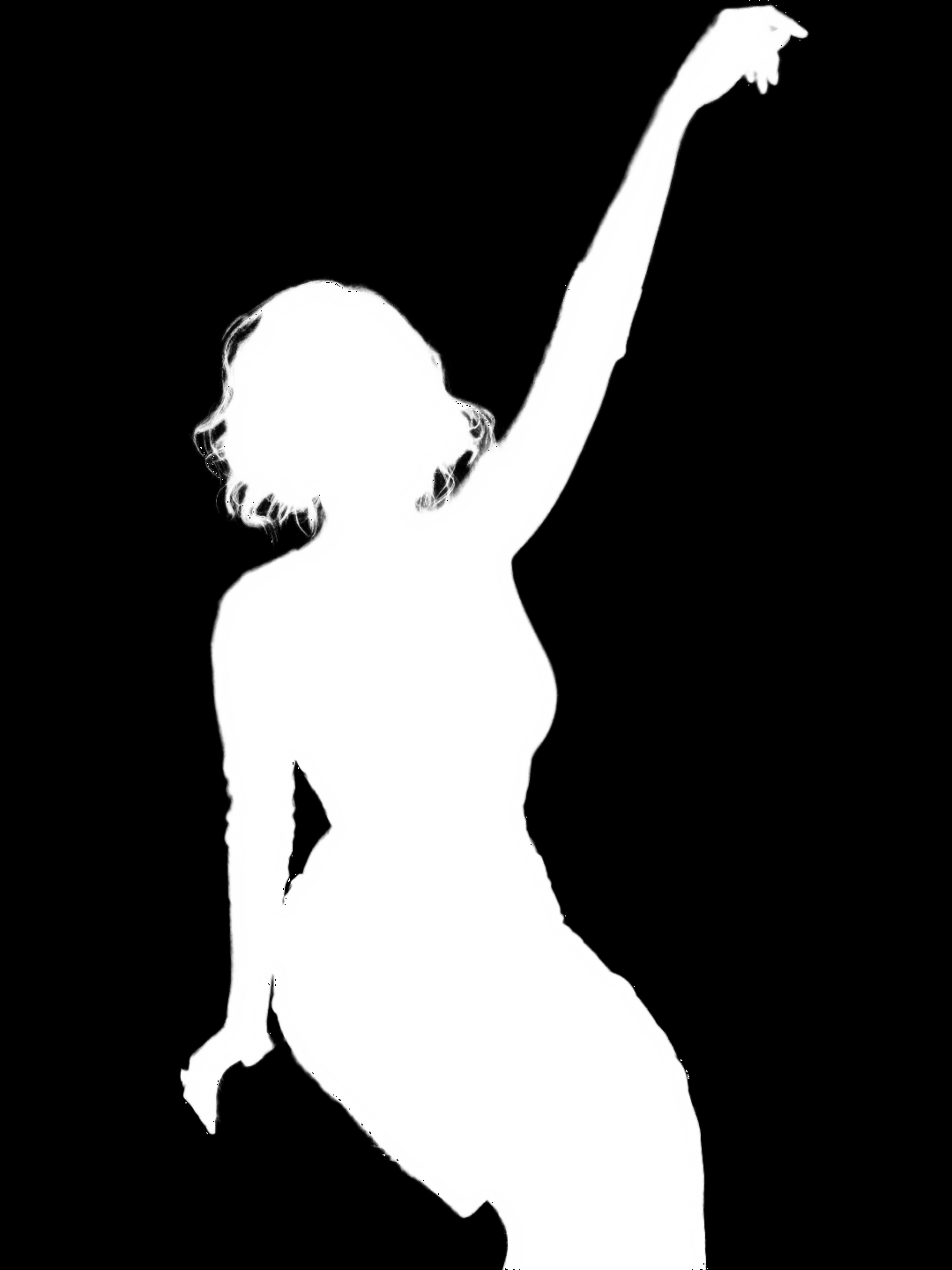}}
\subfloat[Our P3M-Net]{\includegraphics[width=.125\linewidth]{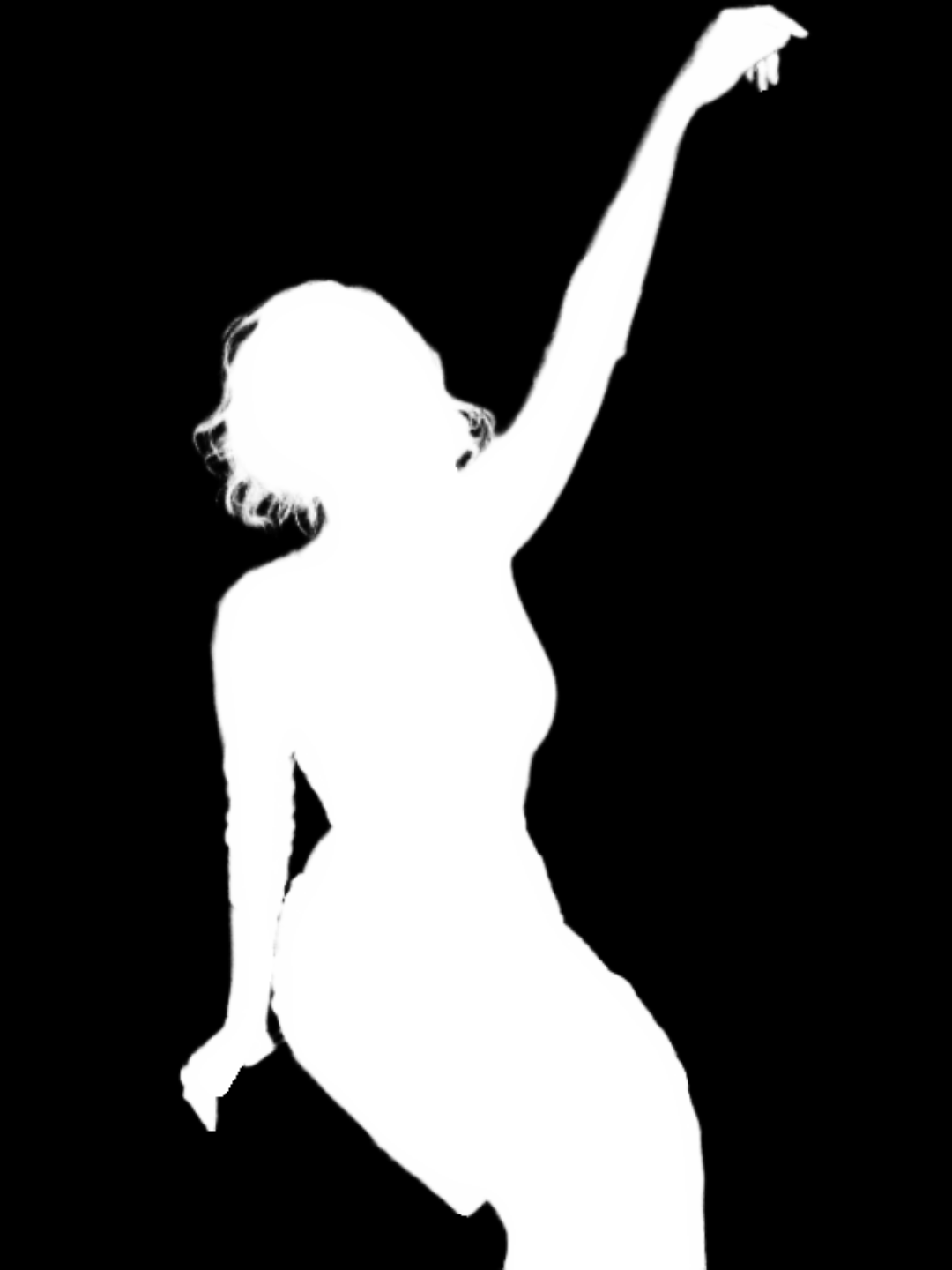}}\\

\caption{Subjective results of different methods on P3M-500-P and P3M-500-NP. Please zoom in for more details.}
\label{fig:experiment}
\end{figure*}

\section{Experiments}
\subsection{Experiment Settings}
To compare the proposed P3M-Net with existing trimap-free methods~\cite{shm,lf,hatt,gfm}, we train them on the P3M-10k face-blurred images and evaluate them on 1) P3M-500-P face-blurred validation set; and 2) P3M-500-NP normal, following the PPT setting.

\noindent\textbf{Implementation Details}
For training P3M-Net, we crop a patch from the image with a size randomly chosen from $512\times512$, $768\times768$, $1024\times1024$, and then resize it to $512\times512$. We randomly flip the patches for the data augmentation. The learning rate is fixed as $1\times10^{-5}$. We train P3M-Net on a single NVIDIA Tesla V100 GPU with a batch size of 8. P3M-Net is trained 150 epochs for about 2 days. It takes 0.132s to test on an $800\times800$ image. For GFM~\cite{gfm} and LF~\cite{lf}, we use the code provided by the authors. For SHM~\cite{shm}, HATT~\cite{hatt} and DIM~\cite{dim} without codes, we re-implement them.

\noindent\textbf{Evaluation Metrics} We follow previous works and adopt the evaluation metrics including the sum of absolute differences (SAD), mean squared error (MSE), mean absolute difference (MAD), gradient (Grad.) and Connectivity (Conn.)~\cite{rhemann2009perceptually}. We calculated them over the whole image for trimap-free methods. We also report the SAD-T, MSE-T, MAD-T metrics within the transition area.

\subsection{Objective and Subjective Results}
The objective and subjective results of different methods are listed in Table~\ref{tab:experiment} and Figure~\ref{fig:experiment}. As can be seen, P3M-Net outperforms all the trimap-free methods in all metrics and even achieves competitive results with trimap-based method DIM~\cite{dim}, which requires the ground truth trimap as an auxiliary input, denotes as DIM$\star$. These results support the designed integration modules, which are able to model abundant interactions between encoder and decoders. As for SHM~\cite{shm}, it has worse SAD than P3M-Net on both datasets, $i.e.$, 21.56 vs. 8.73 and 20.77 vs. 11.23, due to its stage-wise structure, which produces many segmentation errors. LF~\cite{lf} and HATT~\cite{hatt} have large error in transition area, $e.g.$, 12.43 and 11.03 SAD vs. 6.89 SAD of ours, since they lack explicit semantic guidance for the matting task. As in Figure~\ref{fig:experiment}, they have ambiguous segmentation results and inaccurate matting details. GFM~\cite{gfm} is able to predict more accurate semantic mask owing to its multi-task framework. However, it still fails to predict correct context (the last row) and has worse performance than ours, $i.e.$, 13.20 vs. 8.73 in SAD, since it lacks of interactions between encoder and decoders. DIM~\cite{dim} has lower SAD compare with us since it uses ground truth trimap. Nevertheless, P3M-Net still achieves competitive performance in the transition area, $e.g.$, 6.89 vs. 4.89 SAD. It is also noteworthy that even trained only on privacy-preserving training set, most methods can generalize well on arbitrary images, clearly validating the effectiveness of the proposed P3M-10k and the practical value of the PPT setting. Meanwhile, the performance gap when testing on face-blurred images and normal images, $e.g.$, 8.73 SAD vs. 11.23 SAD of P3M-Net, also implies more efforts can be made to advance the research for privacy-preserving portrait matting.

\subsection{Ablation Studies}
We conduct ablation studies of P3M-Net on two datasets P3M-500-P and P3M-500-NP. As seen from Table~\ref{tab:ablation}, the basic multi-task structure without any advanced modules can achieve a fairly good result compared with previous methods \cite{shm,hatt,lf}. With TFI, SAD decreases dramatically to 11.32 and 13.7, owing to the valuable semantic features from encoder and segmentation decoder for matting. Besides, sBFI (dBFI) decreases SAD from 11.32 to 9.47 (9.76) on P3M-500-P and from 13.7 to 12.36 (12.45) on P3M-500-NP, confirming their values in providing useful guidance from relevant visual features. With all three modules, the SAD decreases from 15.13 to 8.73, and 17.01 to 11.23, indicating that our proposed modules bring about 50\% relative performance improvement.

\begin{table}[htb]
\begin{center}
\resizebox{\linewidth}{!}{
\begin{tabular}{ccc|ccc|ccc}
\hline
\multicolumn{3}{c}{} & \multicolumn{3}{|c}{P3M-500-P} & \multicolumn{3}{|c}{P3M-500-NP} \\
\hline
TFI & sBFI & dBFI  & SAD & MSE & MAD & SAD & MSE & MAD\\
\hline
 &  &  & 15.13 & 0.0058 & 0.0088 & 17.01 &0.0062  &0.0099\\
  $\checkmark$& &   & 11.32 & 0.0042 & 0.00066 & 13.7 & 0.0052 &0.008 \\
 $\checkmark$& $\checkmark$ &  & 9.47  & 0.0030 & 0.0055 &12.36  &0.0043  &0.0072  \\
 $\checkmark$&  & $\checkmark$  &9.76  &0.0031  & 0.0057 & 12.45 & 0.0043 &0.0073 \\
 \hline
 $\checkmark$&$\checkmark$  &$\checkmark$  & \textbf{8.73} &\textbf{0.0026}  &\textbf{0.0051}  & \textbf{11.23} & \textbf{0.0035} & \textbf{0.0065}\\
 \hline
\end{tabular}}
\end{center}
\caption{Ablation study of P3M-Net.}
\label{tab:ablation}
\end{table}

\section{Conclusions}
In this paper, we make the first study on the privacy-preserving portrait matting (P3M) problem to respond to the increasing privacy concerns. Specifically, we define the privacy-preserving training (PPT) setting, and establish the first large-scale anonymized portrait dataset P3M-10k, containing 10,000 face-blurred images and ground truth alpha mattes. We empirically find that the PPT setting has little side impact on trimap-based methods while trimap-free methods perform differently, depending on their model structures. We identify that trimap-free methods using a multi-task framework that explicitly models and optimizes both segmentation and matting tasks can effectively mitigate the side impact of PPT. Accordingly, we provide a strong baseline model named P3M-Net, which specifically focuses on modeling the interactions between encoder and decoders, showing promising performance and outperforming all previous trimap-free methods. We hope this study can open a new perspective for the research of portrait matting and attract more attention from the community to address the privacy concerns. 

\bibliographystyle{ACM-Reference-Format}
\bibliography{matting.bib}
\end{document}